\documentclass[10pt]{article}
\usepackage[preprint]{tmlr}

\usepackage{ragged2e}
\usepackage{xcolor}
\usepackage{colortbl}
\usepackage{url}
\usepackage{booktabs}
\usepackage{multirow}
\usepackage{array}
\usepackage{caption}
\usepackage{wrapfig}
\usepackage{graphicx}
\usepackage{algorithm}
\usepackage{algpseudocode}
\usepackage{amsmath}

\usepackage{amssymb}
\usepackage{float}

\usepackage[
  colorlinks=true,
  linkcolor=tmlrBlueDark,
  citecolor=tmlrCiteBlue,
  urlcolor=tmlrBlueDark
]{hyperref}

\usepackage{amsmath,amsfonts,bm}

\def\eqref#1{equation~\ref{#1}}
\def\plaineqref#1{\ref{#1}}
\def\1{\bm{1}}

\DeclareMathAlphabet{\mathsfit}{\encodingdefault}{\sfdefault}{m}{sl}
\SetMathAlphabet{\mathsfit}{bold}{\encodingdefault}{\sfdefault}{bx}{n}

\newcommand{\method}{Random Attention}
\newcommand{\vase}{VaSE}
\newcommand{\snapkv}{SnapKV}
\newcommand{\rkv}{R-KV}
\newcommand{\triattn}{TriAttention}
\newcommand{\dense}{Full}

\newcommand{\mathfive}{MATH500}
\newcommand{\gpqa}{GPQA-D}
\newcommand{\aime}{AIME}
\newcommand{\hmmt}{HMMT}
\newcommand{\lcb}{LiveCodeBench}

\newcommand{\best}[1]{\textbf{#1}}
\newcommand{\sigbelow}[1]{\textcolor{black!50}{#1}}

\newcommand{\budget}{K}                  % scored KV budget per head
\newcommand{\residual}{r}                % recent-buffer length (64)
\newcommand{\promptlen}{\ell_{\mathrm{p}}} % prompt (question) length
\definecolor{oursbg}{RGB}{253,236,233} % light tint of the figures' ours-red for the \method row
\definecolor{oursred}{RGB}{227,73,72}
\newcommand{\gain}[1]{{\scriptsize\textcolor{oursred}{#1}}}
\newcommand{\gainnull}[1]{{\scriptsize\textcolor{black!45}{#1}}}

\title{Random Attention: Rethinking KV Cache Eviction for Efficient Reasoning}

\long\def\paperabstract{Large language models achieve superior performance on tasks that require extended reasoning, but long chains of thought make the KV cache a severe memory bottleneck. Existing KV cache compression methods share one paradigm: score each cached token by some estimate of how much it will matter later, and keep the top-scoring ones. We show that the selection signal contributes almost nothing. \emph{Random Attention} keeps the prompt and evicts uniformly at random within each attention head, computing no score at all; across four models and six reasoning tasks, it matches the strongest baseline in task performance while delivering $32$--$43\%$ higher throughput than that method when deployed with vLLM. Controlled experiments explain this by showing that 1) the prompt is the fragile part of the cache, and most of the gap between selectors is just whether their selection signal happened to keep it; 2) the reasoning trace protects itself against eviction with redundancy at two levels, in the text (the model restates what it still needs as it works) and across attention heads (each keeps its own copy of the trace), so once the prompt is safe, a random draw retains enough copies of what the model still needs, and no score is required to pick them. Our code is publicly available at \url{https://github.com/SalesforceAIResearch/Random-Attention}.}

\author{Heng Wang$^{1,2}$, Jielin Qiu$^{1}$, Wenting Zhao$^{1}$, Cheng Qian$^{1,2}$, Liangwei Yang$^{1}$, Weizhi Zhang$^{1,3}$,\\
Jiawei Han$^{2}$, Heng Ji$^{2}$, Silvio Savarese$^{1}$, Shelby Heinecke$^{1}$, Huan Wang$^{1}$\vspace{1.5mm}\\
$^1$Salesforce AI Research, $^2$University of Illinois Urbana-Champaign, $^3$ University of Illinois Chicago
 \vspace{1mm}\\
}

\def\month{09}
\def\year{2026}

\begin{document}

\maketitle

\section{Introduction}

\begin{figure}[!h]
\centering
\includegraphics[width=\textwidth]{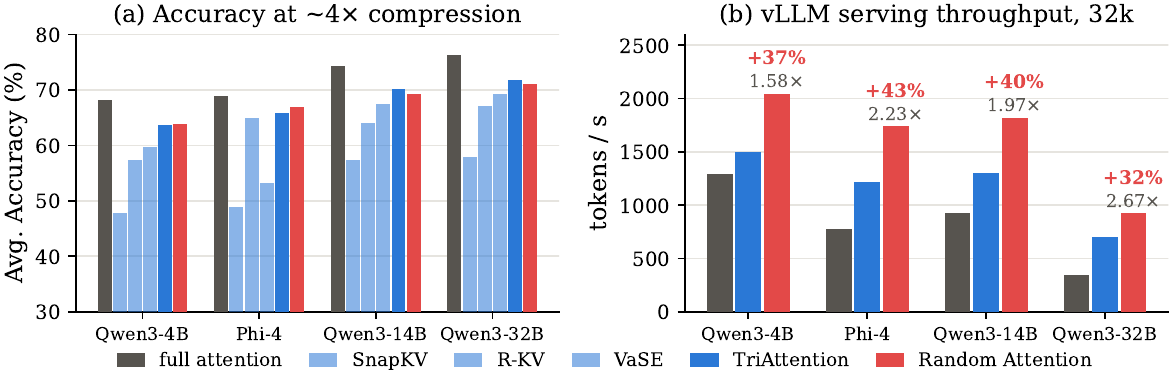}
\caption{\textbf{(a)} Mean accuracy over the six reasoning tasks of
Tables~\ref{tab:main} and~\ref{tab:appendix_grid} at ${\sim}4\times$
compression: \method{} matches the strongest baseline on every model
(the small gaps to \triattn{} at 14B and 32B are mostly driven by code
reasoning, where long prompts consume the budget, \S\ref{sec:main}).
\textbf{(b)} vLLM serving throughput at $32$k-token generations
(Table~\ref{tab:vllm}); labels give \method{}'s multiple of full
attention and its margin over \triattn{}: with no scoring pass it
serves $32$--$43\%$ higher throughput.}
\label{fig:overview}
\end{figure}

Reasoning models \citep{deepseekr1, openai2024o1, phi4reasoning} solve hard problems by generating chains of thought that
run to tens of thousands of tokens. The key-value (KV) cache grows linearly with the length of generation, creating a severe memory bottleneck. KV cache eviction methods have been developed to address this by
keeping a fixed budget of KV cache entries and discarding the rest as
decoding proceeds. Existing KV cache eviction methods follow the same paradigm: score each cached token by an estimate of how much it will matter later and keep the top-scoring ones. This line of work is a sequence of better
scores, from accumulated attention \citep{zhang2023ho}, attention from a recent window \citep{li2024snapkv}, and attention combined with redundancy
\citep{cai2025rkv}, to value magnitude \citep{chang2026valueaware} and position-dependent key
statistics \citep{mao2026triattention}. Each new score is motivated by heuristics observed to be correlated with task accuracy, and the premise behind all of them is that the score decides accuracy under compression.

We test that premise directly and show that the selection signal contributes almost nothing. \emph{Random Attention} keeps the prompt
and evicts uniformly at random within each attention head, computing
no score at all. Across four models (Qwen3-4B, 14B and 32B, and
Phi-4-reasoning) and six reasoning tasks spanning math, science and
code, it is comparable to the strongest baseline (Figure~\ref{fig:overview}a) and even significantly ahead in 31 of the 60 baseline comparisons in the main result table. Served through vLLM integration, it delivers $32$--$43\%$ more tokens per second at $32$k-token generations than the strongest baseline, since it never runs a scoring pass (Figure~\ref{fig:overview}b). The only cell of that table where a baseline is significantly ahead is code reasoning on the largest Qwen3 model, traced in \S\ref{sec:main} to long prompts that shrink the budget left for the trace. The selection signal, in other words, contributes almost nothing beyond uniform random selection.

Two controlled experiments explain why. First, \textbf{the prompt is the fragile part of the cache}. Prior evictors differ in whether the
prompt survives, some pinning it by rule and others leaving it to the
score, so once every method is given the same rule (keep the prompt),
most of the gap between them disappears, and each method's gain is ordered by how much of the question its score has been losing
(\S\ref{sec:mech_protect}). Second, \textbf{the reasoning trace protects
itself}. It is stored redundantly at two levels, in the text, because
the model restates what it is still using, and across attention heads,
because every head holds its own copy of every token and eviction
decides per head which copies die. A planted-fact probe shows the model
reading a value from whichever heads still hold it: a fact kept in one
head is almost never retrieved, kept in several it almost always is, and
the shape of the surviving copies does not matter
(\S\ref{sec:mech_pool}). Once the prompt is safe, a random draw
keeps enough copies of what the model still needs; what a signal still buys is the rare fact
stated once and never restated, which reasoning traces seldom produce
(\S\ref{sec:mech_boundary}).

Our findings have two practical implications. First, Random Attention
is a deployable method in its own right. It needs no calibration, no
tuning and no scoring pass, and it is the fastest evictor we measured
at equal accuracy, so it is a reasonable default for serving reasoning
models under a memory budget, and the baseline that any new selection
signal has to beat at matched budget and matched prompt protection.
Second, the findings redirect what eviction research should optimise.
The accuracy of an evictor is decided by what it protects, not by how it ranks the rest, which moves the open questions to where protection still matters: how to budget long prompts, especially in code tasks where protecting the entire prompt consumes a substantial fraction of the cache budget, and how to recover rare once-stated facts that only a content-dependent signal can preserve (\S\ref{sec:mech_boundary}).

\section{Preliminaries}
\label{sec:setting}

\paragraph{Setting.}
We study KV cache \emph{eviction during decoding}. This is the regime reasoning
models create: a model answering a short (e.g.\ $200$-token) math question may generate a very long chain of thought (e.g.\ more than $10{,}000$ tokens). Eviction permanently discards key-value pairs
once the cache reaches a budget, which bounds memory but risks irrecoverable
loss; it is therefore distinct from sparse-attention \emph{selection}, which
attends to a subset but keeps every pair in memory and so still grows linearly
with sequence length. Everything below concerns eviction.

\paragraph{Notation.}
Let $t$ index decode steps. At step $t$ an attention head holds $N$
cached key-value pairs $(k_i, v_i)$, $i = 1,\dots,N$, with
$k_i, v_i \in \mathbb{R}^{d}$, where $i$ is the position of the token
that produced the pair; positions $1,\dots,\promptlen$ are the prompt.
The head's output for the current query $q_t$ is $o_t=\sum_{i\le N}\alpha^{(t)}_i v_i$, with attention weights $\alpha^{(t)}_i=\mathrm{softmax}_i\big(q_t^{\top}k_i/\sqrt{d}\big)$. We write $v_{ij}$ for the $j$-th coordinate of $v_i$ and $\|k_i\|$ for
the norm of $k_i$. Eviction decisions are made independently in every layer and KV head,
as is standard; we drop both indices throughout.

\paragraph{Periodic eviction with budget $\budget$ and buffer $\residual$.}
We adopt the decode-phase framework of \citet{cai2025rkv} and \citet{chang2026valueaware}. The cache
keeps a persistent budget of $\budget$ pairs plus a buffer of the
$\residual \ll \budget$ most recent pairs, which is never scored. Each decode
step appends one pair, so the buffer fills every $\residual$ steps and triggers
an eviction: every candidate $i$ in the candidate set $\mathcal{C}_t$,
the cached positions outside the buffer, receives a real-valued score
$s_i$ from a policy-specific rule, which may depend on the whole cache
and on the past queries, and the $\budget$ highest-scoring candidates
are kept,
\begin{equation}
\label{eq:keep}
\mathcal{S}_t \;=\; \operatorname{top-}\!\budget_{\,i \in \mathcal{C}_t}\; s_i ,
\end{equation}
returning the cache to $\budget + \residual$ entries per head. Eviction is monotonic: a discarded pair is gone for good, as in a memory-bounded
deployment. Eviction methods differ mostly in $s$.

\paragraph{Baselines as choices of $s$.}
We write each prior method's $s$ in the notation above,
with every score evaluated at the eviction step $t$.
\textbf{StreamingLLM} \citep{xiao2024efficient} has no score: it keeps
the first few (attention-sink) positions and the recent buffer. \textbf{H2O} \citep{zhang2023ho} keeps the
positions that have received the most attention since they entered the
cache, $s_i=\sum_{t'=i}^{t}\alpha^{(t')}_i$. \textbf{SnapKV}
\citep{li2024snapkv} uses only the last $w$ queries,
$s_i=\sum_{t'=t-w+1}^{t}\alpha^{(t')}_i$, max-pooled over neighbouring
positions.   \textbf{R-KV} \citep{cai2025rkv} mixes the SnapKV score with a redundancy penalty, $s_i = \lambda\, s^{\mathrm{Snap}}_i - (1-\lambda)\,\mathrm{softmax}_i(\bar{c})$, where $\bar{c}_i$ is the mean cosine similarity of $k_i$ to the other cached keys, not counting its most recent near-duplicates, so that a restated fact is not kept twice. \textbf{VaSE} \citep{chang2026valueaware} scores values rather than keys: it
keeps the $n_{\mathrm{large}} < \budget$ positions with the
largest value range $\max_{j} v_{ij}-\min_{j} v_{ij}$ and fills the
remaining $\budget - n_{\mathrm{large}}$ slots by sampling positions
with probability proportional to their SnapKV score.
\textbf{TriAttention} \citep{mao2026triattention} predicts the attention a key will receive from a per-head query centre calibrated offline: $s_i$ averages $f_{k_i}(t-i+\delta)$ over future offsets $\delta \in \{1,2,4,\dots,2^{16}\}$, where $f_{k_i}$ is a trigonometric series in the query--key distance whose coefficients depend on the centre and on $k_i$'s pre-RoPE components, plus a norm term in $\|k_i\|$.

\section{Random Attention}
\label{sec:randomattention}

   \begin{figure}[t]
   \centering
   \includegraphics[width=\textwidth]{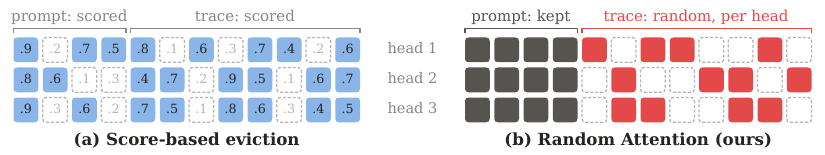}
   \caption{Overview of the methods. \textbf{(a)} Baselines keep $\budget$ positions per head by a score recomputed at every eviction; except in \triattn{}, the prompt competes too. \textbf{(b)} \method{} pins the prompt and samples the rest uniformly, independently per head. Dashed cells are evicted.}
   \label{fig:method}
   \end{figure}
   
As illustrated in Figure~\ref{fig:method}, \method{} is a signal-free eviction policy defined by two structural choices and nothing else. The intuition is to separate the irreplaceable input from the model-generated trace: the question is stated once and cannot be recovered if evicted, whereas the trace revisits and restates its intermediates as generation proceeds. We protect the former and use random sampling for the latter:

\begin{enumerate}
\item \textbf{Protect the question.} Positions $1,\dots,\promptlen$,
  the entire prefill (the system prompt, chat template, and question),
  are never evicted.
\item \textbf{Scatter the rest, per head.} Every remaining cached position receives
  an i.i.d.\ uniform random score, and each KV head keeps its top-$\budget$
  independently. Sampling without a signal spreads the retained budget evenly over
  the whole trace, and differently in every head.
\end{enumerate}

In the notation of \S\ref{sec:setting}, $s_i{=}{+}\infty$ for $i \le \promptlen$ and $s_i \sim \mathrm{Uniform}(0,1)$ otherwise, drawn independently per KV head at every eviction; Eq.~\plaineqref{eq:keep} does the rest, so an eviction costs one $\mathrm{rand}$ and one $\operatorname{topk}$. This is the weakest
selection signal we can write down: \method{} is both a deployable method and a
null hypothesis. Any signal-based selector that cannot beat it at matched budget is not
extracting usable information from its signal.

% \paragraph{Why this could work.}
% Two properties make the scatter a priori defensible. Protecting the prompt
% removes the one catastrophic, structurally identifiable loss without computing
% any score. And a uniform per-head draw gives every remaining position a
% nonzero survival probability in every head, whereas any deterministic spend of
% the same budget --- a longer recency window, or top-$k$ under any score ---
% keeps some positions with probability zero and loses whatever falls there with
% certainty. Randomization has precedent elsewhere in KV caching
% \citep{rlt,kvec}, but plausibility is not an explanation for the measured
% result --- that the scatter \emph{ties or beats} learned selection on real
% reasoning tasks (\S\ref{sec:experiments}). \S\ref{sec:mechanism} supplies the
% explanation with pre-registered causal interventions on cache storage:
% reasoning traces redundantly restate load-bearing facts verbatim, a fact's
% storage pools across a few carrier KV heads down to single-token granularity, and
% retrieval depends on how many of those carrier copies survive --- a quantity a
% per-head uniform scatter preserves better than any attention-derived ranking.

\section{Experiments}
\label{sec:experiments}

\subsection{Setup}
\label{sec:eval_protocol}

\paragraph{Baselines.}
We compare against the four eviction methods of \S\ref{sec:setting},
\snapkv{}, \rkv{}, \vase{} and \triattn{}, each run as released at
the same budget (detailed configurations in Appendix~\ref{sec:engine}). All four are training-free and can run on any model as
released, which is the comparison our question calls for. Learned eviction policies \citep{bui2026make, bui2026cache, dong2026foresightkv}, which
train a scorer for each base model, are discussed in \S\ref{sec:related}. \textbf{Full} attention (no eviction) is the ceiling.
The diagnostics of \S\ref{sec:mech_protect} additionally use \emph{recency+prompt}
(keep the KV caches corresponding to the prompt, fill with the contiguous recent window,
StreamingLLM-style \citep{xiao2024efficient}). All methods are evaluated with FlashAttention-2 kernels \citep{dao2024flashattention}, without PagedAttention \citep{kwon2023efficient}.

\paragraph{Models and tasks.}
We evaluate Qwen3-4B, Qwen3-14B, Qwen3-32B \citep{qwen3}, and Phi-4-reasoning (14B)
\citep{phi4reasoning} across six reasoning tasks spanning math, science, and code: \mathfive{} \citep{hendrycks2021measuring,lightman2024lets} ($500$ problems), GPQA-Diamond (\gpqa{}) \citep{rein2024gpqa} ($198$ problems), AIME 2025 and 2026
($30$ problems each, reported as one pooled \aime{} column) and \hmmt{}
($60$ problems) via MathArena \citep{balunovic2026matharena}, and \lcb{}-v6 medium
\citep{jain2025livecodebench} ($383$ problems; pass@1 by real test execution).
Generation uses each model's released sampling settings
(temperature $0.6$ for the Qwen3 models, $0.8$ for Phi-4-reasoning;
nucleus $p{=}0.95$ for all), and each result is repeated as independently sampled runs: $2$ on \mathfive{}, $4$ on \gpqa{} and \lcb{}, and
$16$ on \aime{} and \hmmt{}. Every accuracy in the paper is the average over those repeated runs. The main grid fixes each task's budget at ${\sim}4\times$ compression of its
typical trace (${\sim}3\times$ for \lcb{}); the per-head budget
$\budget$ for each task appears in Table~\ref{tab:main}'s header and is
detailed in Appendix~\ref{sec:engine}. We set the maximum generation length to 32,768 (32k) tokens.

\paragraph{Metrics and statistics.}
The primary metric is accuracy, judged by whether the final boxed answer is correct (following \cite{chang2026valueaware} and \citet{gao2026sparse}). Every claimed margin is gated by a paired, problem-clustered percentile
bootstrap ($95\%$ CI) plus an exact sign test; grid cells significantly below \method{} are grayed.

\subsection{Main Results}
\label{sec:main}

Table~\ref{tab:main} presents the performance on Qwen3-4B, Phi-4-reasoning, and Qwen3-32B; Qwen3-14B replicates the pattern at an intermediate scale in Appendix~\ref{sec:generality}. Paired tests put \method{}
significantly ahead in $31$ of the table's $60$ baseline cells and
significantly behind in one.

% AUTO-GENERATED by kvcompress/eval/gen_paper_tables.py -- do not hand-edit.

\begin{table}[t]
\centering
\footnotesize
\setlength{\tabcolsep}{3.5pt}
\caption{Accuracy under KV cache eviction at each task's ${\sim}4\times$ compression (\lcb{}: ${\sim}3\times$); the header gives each task's per-head KV budget $\budget$ (\S\ref{sec:setting}). \best{Bold}: best eviction method per column; \sigbelow{grayed}: significantly below \method{} (paired clustered bootstrap + sign test, 95\%).}
\label{tab:main}
\resizebox{0.88\textwidth}{!}{%
\begin{tabular}{lccccc}
\toprule
 & \mathfive{} & \gpqa{} & \aime{} & \hmmt{} & \lcb{} \\
 & \scriptsize $K{=}1024$ & \scriptsize $K{=}2048$ & \scriptsize $K{=}4096$ & \scriptsize $K{=}4096$ & \scriptsize $K{=}3072$ \\
\midrule
\rowcolor{black!7}
\multicolumn{6}{l}{\textbf{Qwen3-4B}} \\
\dense{} & 0.939 & 0.562 & 0.642 & 0.462 & 0.807 \\
\noalign{\vskip -2pt}
\cmidrule(lr){1-6}
\snapkv{} & \sigbelow{0.703} & \sigbelow{0.369} & \sigbelow{0.418} & \sigbelow{0.395} & \sigbelow{0.507} \\
\rkv{} & \sigbelow{0.810} & \sigbelow{0.482} & \sigbelow{0.494} & \sigbelow{0.371} & \sigbelow{0.712} \\
\vase{} & \sigbelow{0.809} & \sigbelow{0.461} & 0.596 & 0.421 & \sigbelow{0.700} \\
\triattn{} & 0.864 & \best{0.533} & 0.592 & 0.437 & \best{0.755} \\
\rowcolor{oursbg}
\method{} (ours) & \best{0.874} & 0.530 & \best{0.610} & \best{0.438} & 0.744 \\
\midrule
\rowcolor{black!7}
\multicolumn{6}{l}{\textbf{Phi-4-reasoning}} \\
\dense{} & 0.922 & 0.707 & 0.677 & 0.444 & 0.697 \\
\noalign{\vskip -2pt}
\cmidrule(lr){1-6}
\snapkv{} & \sigbelow{0.844} & \sigbelow{0.442} & \sigbelow{0.502} & \sigbelow{0.343} & \sigbelow{0.314} \\
\rkv{} & 0.909 & 0.636 & 0.643 & \best{0.440} & \sigbelow{0.621} \\
\vase{} & \sigbelow{0.853} & \sigbelow{0.562} & \sigbelow{0.520} & \sigbelow{0.354} & \sigbelow{0.373} \\
\triattn{} & 0.891 & \best{0.684} & 0.633 & 0.431 & 0.652 \\
\rowcolor{oursbg}
\method{} (ours) & \best{0.910} & 0.678 & \best{0.662} & 0.430 & \best{0.667} \\
\midrule
\rowcolor{black!7}
\multicolumn{6}{l}{\textbf{Qwen3-32B}} \\
\dense{} & 0.950 & 0.703 & 0.715 & 0.559 & 0.886 \\
\noalign{\vskip -2pt}
\cmidrule(lr){1-6}
\snapkv{} & \sigbelow{0.816} & \sigbelow{0.476} & \sigbelow{0.541} & 0.450 & \sigbelow{0.609} \\
\rkv{} & 0.857 & 0.638 & \sigbelow{0.613} & 0.472 & 0.779 \\
\vase{} & \sigbelow{0.868} & \sigbelow{0.597} & \best{0.680} & \best{0.524} & 0.797 \\
\triattn{} & 0.887 & \best{0.683} & 0.677 & 0.508 & \best{0.834} \\
\rowcolor{oursbg}
\method{} (ours) & \best{0.891} & \best{0.683} & 0.664 & 0.509 & 0.806 \\
\bottomrule
\end{tabular}}
\end{table}

\paragraph{A selection signal buys nothing on math and science reasoning.}
On \mathfive{} and \gpqa{}, \method{} beats \vase{} and \snapkv{}
significantly on every model, and \rkv{} significantly on Qwen3-4B. No selector
beats it significantly on these tasks: the leads that do appear
(\triattn{} by $0.3$--$0.6$ points on two \gpqa{} cells) sit inside the noise.

\paragraph{On competition math no selector pulls ahead.}
Competition math tasks including \aime{} and \hmmt{} make the
comparison ride on a smaller and harder sample. With $16$ sampled runs per problem,
\snapkv{} still trails \method{} significantly on every model, as do
\rkv{} on the Qwen3 models and \vase{} on Phi-4-reasoning; no selector
in Table~\ref{tab:main} is significantly above \method{} on either
task, and the nominal leads
run both ways: \vase{} edges it on Qwen3-32B by 1.6 and 1.5 points, against a run-to-run standard deviation of about 4 points for both methods on 60-problem sets. These tasks separate only once the budget
tightens (\S\ref{sec:regime}), and then in \method{}'s favour.

\paragraph{On code reasoning most signal-based selectors fall apart due to much longer prompts.}
\lcb{} is the one task with large gaps: \snapkv{} loses $20$--$35$ points to \method{} on every model, \vase{} collapses on
Phi-4-reasoning ($0.373$, $29$ points behind) and is grayed on two of the three models, as is \rkv{}. \triattn{} and \method{} tie on Qwen3-4B and Phi-4-reasoning, while \triattn{} leads on Qwen3-32B by about three points, the single significant baseline win in the main grid. The prompt is what sets code apart. \lcb{} prompts average $557$ tokens, which is six times \mathfive{}'s under the same tokenizer, and the longest can consume up to half of the $\budget{=}3072$ budget. Therefore a selector that fails to capture the prompt loses more here than anywhere else; we show in the following section that protecting it closes the \snapkv{} and \vase{} gaps (\S\ref{sec:mech_protect}). \method{} itself
pins every prompt token, so on code reasoning a large, variable share
of its budget is spent before selection begins; much of a code prompt is
scaffolding (I/O formats, harness instructions) that a smarter rule
might compress rather than pin whole, which we leave to future work since \method{}'s value as a null lies in having nothing to tune.

\begin{figure}[t]
\centering
\includegraphics[width=0.95\textwidth]{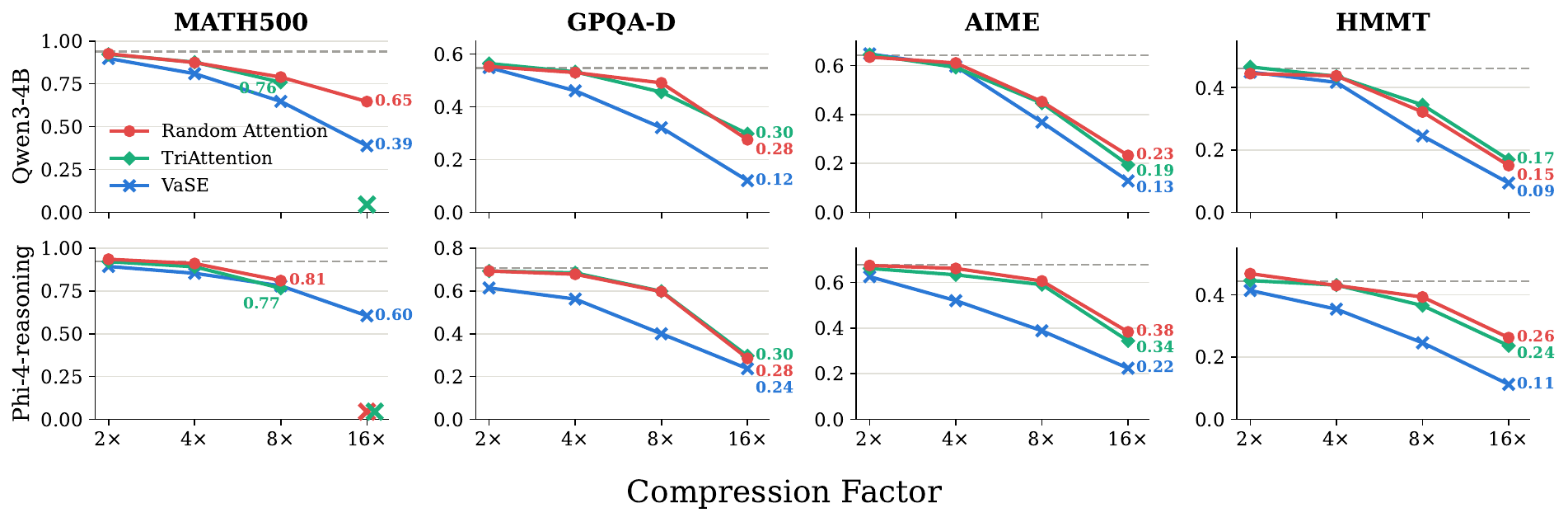}
\caption{Accuracy from $2\times$ to $16\times$ compression on Qwen3-4B
and Phi-4-reasoning, on the four math and science tasks; dashed lines mark full attention.}
\label{fig:regime}
\end{figure}

\subsection{Compression Pressure Widens the Gap, in Every Family}
\label{sec:regime}

Figure~\ref{fig:regime} presents the performance from $2\times$ to $16\times$ compression on Qwen3-4B and Phi-4-reasoning, on the four math and science tasks, and both families tell the same story: at $2\times$ every method sits near full attention; as the budget tightens, \method{} stays tied with \triattn{}, while the gap from both to \vase{} opens. \lcb{} is left out of the sweep since its prompts alone would not fit the small budgets.

\section{Why the Selection Signal Buys So Little}
\label{sec:mechanism}
The content of the KV cache in reasoning can be divided into two types. The prompt is stated once
and never stated again. The working state (i.e., the intermediate reasoning steps a solution builds on) is written and rewritten continually as the model
reasons. We show that the first is fragile, and methods differ in how they treat it; the second is redundant enough that a random draw over it keeps what the model still needs.

% AUTO-GENERATED by kvcompress/eval/gen_paper_tables.py -- do not hand-edit.

\begin{table}[t]
\centering
\small
\setlength{\tabcolsep}{4pt}
\caption{Performance before and after protecting the prompt. \rkv{} never gains more than $1.9$ points, \snapkv{} gains everywhere, \vase{} gains materially only on Phi-4-reasoning.}
\vspace{-2mm}
\label{tab:promptprotect}
\resizebox{0.95\textwidth}{!}{%
\begin{tabular}{lcccccccc}
\toprule
 & \multicolumn{4}{c}{Qwen3-4B} & \multicolumn{4}{c}{Phi-4-reasoning} \\
\cmidrule(lr){2-5} \cmidrule(lr){6-9}
 & \multicolumn{2}{c}{\mathfive{}} & \multicolumn{2}{c}{\gpqa{}} & \multicolumn{2}{c}{\mathfive{}} & \multicolumn{2}{c}{\gpqa{}} \\
\cmidrule(lr){2-3} \cmidrule(lr){4-5} \cmidrule(lr){6-7} \cmidrule(lr){8-9}
Method & score alone & $+$ prompt & score alone & $+$ prompt & score alone & $+$ prompt & score alone & $+$ prompt \\
\midrule
\snapkv{} & 0.703 & 0.829\,\gain{+12.6} & 0.369 & 0.492\,\gain{+12.3} & 0.844 & 0.889\,\gain{+4.5} & 0.442 & 0.667\,\gain{+22.5} \\
\rkv{} & 0.810 & 0.812\,\gainnull{+0.2} & 0.482 & 0.471\,\gainnull{-1.1} & 0.909 & 0.902\,\gainnull{-0.7} & 0.636 & 0.655\,\gainnull{+1.9} \\
\vase{} & 0.809 & 0.812\,\gainnull{+0.3} & 0.461 & 0.470\,\gainnull{+0.9} & 0.853 & 0.895\,\gain{+4.2} & 0.562 & 0.664\,\gain{+10.2} \\
\midrule
Recency window & 0.246 & 0.843\,\gain{+59.7} & 0.093 & 0.519\,\gain{+42.6} & 0.665 & 0.884\,\gain{+21.9} & 0.323 & 0.658\,\gain{+33.5} \\
\rowcolor{oursbg} \method{} & 0.459 & \best{0.874}\,\gain{+41.5} & 0.231 & \best{0.530}\,\gain{+29.9} & 0.759 & \best{0.910}\,\gain{+15.1} & 0.434 & \best{0.678}\,\gain{+24.4} \\
\bottomrule
\end{tabular}}
\vspace{-3mm}
\end{table}

\subsection{The Prompt Is the Fragile Part}
\label{sec:mech_protect}
\label{sec:mech_why}   % kept: \S\ref{sec:mech_why} is cited from method.tex and experiments.tex

Methods disagree about the prompt. \triattn{} keeps the whole input by default, while \vase{}, \rkv{}, and
\snapkv{} only keep the sink tokens \citep{xiao2024efficient, han-etal-2024-lm} and leave every slot to the score, so a comparison across papers also compares protection regimes \citep{chen-etal-2026-pitfalls}. Giving every method the same rule separates the score from the protection (Table~\ref{tab:promptprotect}), and one pattern orders the outcome: \emph{the rule pays each method according to how much of the question its score was losing}, measured for every selector by logging, round by
round, how much of the prompt it keeps (Appendix~\ref{sec:keeplog}).
\snapkv{}, whose score retains the least of the prompt, always gains,
up to $22.5$ points on Phi-4-reasoning \gpqa{}; \vase{} gains only where its
retention fails, little on Qwen3-4B but $+4.2$ and $+10.2$ on
Phi-4-reasoning, whose prompts are two to three times longer; \rkv{},
which retains the most, never gains more than $1.9$ points (survival
fractions in Appendix~\ref{sec:keeplog}). Once every method keeps the prompt, the three baselines land within $2.2$ points of one another in every setting. On Phi-4-reasoning they also land within about two points of \method{}, which ranks nothing; on Qwen3-4B a residual of $4$--$6$ points below \method{} remains for all three, and \rkv{} and \vase{}, which already kept most of the prompt there, gain almost nothing from the rule. Code reasoning on both models, competition math on Qwen3-4B and \gpqa{} at
32B (Appendix~\ref{sec:ppappendix}) show the same pattern: the rule
closes every gap that was large and method-specific, and what survives
it is smaller and runs in \method{}'s favour. Most of the difference
between the baselines was therefore the prompt. Whatever their scores
add beyond it is small, and where a residual remains it is a deficit:
with the prompt protected, each of the three still trails a policy that ranks nothing.

The two signal-free rows make the same point from the other side.
Without the rule, a recency window allocates all the budget to the recent trace and none to the prompt and scores as low as $0.09$, and \method{}, which
keeps the prompt only at the uniform rate, falls to $0.23$--$0.76$.
With the rule, the same two policies lose nothing that matters:
\method{} is the best policy in every setting and a plain recency
window is never more than two points below the best protected baseline. Losing the prompt
is catastrophic and cutting the trace at random is not, which is what
makes the prompt the fragile part of the cache. The same confound explains the large gaps others report between random
retention and signal-based selection in previous works \citep{semanticshbm,holdthought}:
their random baselines perform poorly since the prompt is lost.

\subsection{The Working State Protects Itself}
\label{sec:mech_pool}
\label{sec:mech_synth}   % alias kept: the planted-fact probe is cited from the appendix

The rest of the cache is the model's own reasoning trace (working state), which is stored redundantly at two levels. The first redundancy is in
the text and is already shown by \citet{cai2025rkv}: reasoning traces restate what they are still using, so a value that matters rarely lives at one position only. The second is across heads: each of the KV heads caches its own copy of every token, and eviction decides per head which copies die. A token is only lost when all KV heads happen to drop it.

We show how the model makes use of the second, cross-head redundancy with a planted-fact probing experiment. A synthetic fact (e.g.\ \texttt{Let zq = 4729}; a fresh variable and value each
time) is inserted into real model-generated \mathfive{} reasoning traces, with a question needing the value appended at the end. The fact is planted $1{,}536$ tokens before the question, so the cache
is evicted $15$ times between the two (distance grid in Appendix B); the question itself is always kept. What
we control is which key-value heads keep the fact: a \emph{condition}
is a chosen set of heads in which the fact's tokens are pinned, with
the fact evicted from every other head and the standard per-head
uniform eviction running on everything else. Two metrics measure what survives: \emph{retrieval}, the fraction of traces whose greedy decode reproduces the value, and, where retrieval floors, a graded \emph{recall} $R$, the share of the log-probability gap between deleting the fact from every cache ($R{=}0$) and keeping it in every cache ($R{=}1$) that the surviving copies recover (Appendix~\ref{sec:synthapp}). Every condition is scored on the same $250$--$500$ planted traces.

\begin{figure}[t]
\centering
\includegraphics[width=\textwidth]{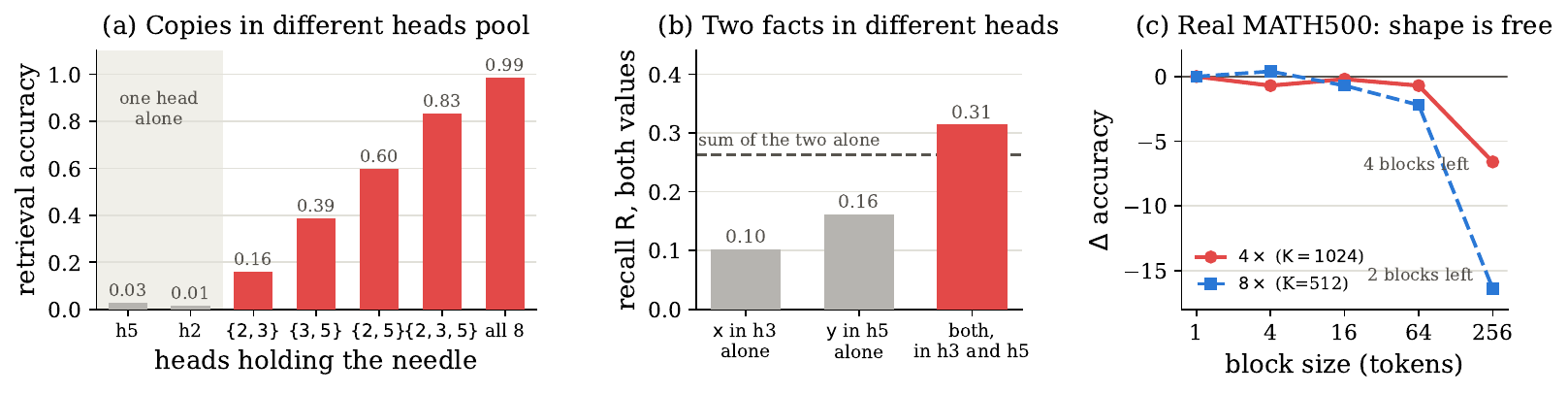}
\vspace{-5mm}
\caption{\textbf{(a)} A fact held in one head is almost never retrieved;
held in several it is. \textbf{(b)} Two facts in \emph{different} heads
are worth more together than the sum of each alone (dashed).
\textbf{(c)} Real \mathfive{}: contiguous blocks cost nothing up to
size $64$; accuracy drops only once a head is left with $4$
($K{=}1024$) or $2$ ($K{=}512$) blocks.}
\label{fig:mechanism}
\vspace{-3mm}
\end{figure}

\paragraph{Copies pool across heads.} Attention heads specialise: only
three of Qwen3-4B's eight key-value heads retain a usable trace of the
fact on their own, consistent with the retrieval-head specialisation of
\citet{wu2025retrieval}, and even those three are weak alone: the
best single head yields the value in $3\%$ of trials, the next in
$1\%$. But the readout does not depend on any one head: the same two
heads together yield the fact in $60\%$ of trials, three heads in
$83\%$, and all eight in $99\%$ (Figure~\ref{fig:mechanism}a). Pooling
is thus strongly superadditive, a pair being worth many times the sum
of its singles, and it even crosses facts: two values held in
\emph{different} heads, both needed by the answer, give $R{=}0.31$
together against $0.10$ and $0.16$ alone
(Figure~\ref{fig:mechanism}b). For an eviction policy the consequence
is direct: a value stays usable as long as \emph{some} heads keep a
copy, which is exactly what independent per-head draws maximise. On
real \mathfive{} traces this extra coverage is not even needed: a
shared draw, the same random positions in every head, scores within
$0.3$ points of \method{} at $4\times$ and $8\times$
(Appendix~\ref{sec:keeplog}), because the text-level redundancy already
keeps a restated copy; the cross-head level carries what the text does
not restate, which is the probe's regime.

\paragraph{The shape of the copies does not matter.} We deal the fact
out token by token across heads, so that no two consecutive tokens
share a head and no head holds a readable span; retrieval barely moves
($0.33$, against $0.39$ for an intact sentence at the same retained
mass) and recall is essentially unchanged ($R{=}0.75$ vs.\ $0.76$).
The same insensitivity appears on real \mathfive{} traces at full
scale: keeping the history in contiguous blocks rather than scattered
tokens costs nothing as blocks grow from $1$ to $64$ tokens; accuracy
drops only at block size $256$, where the budget leaves a head just
four or two blocks, and drops more with two, so what matters is
blocks per head, not block length (Figure~\ref{fig:mechanism}c). Together the two
findings say the
answer depends on whether some usable copy
of a needed value survives \emph{somewhere}, not on which copy, in
which head, or in what shape.

\subsection{What Is Left for a Selection Signal}
\label{sec:mech_boundary}

% AUTO-GENERATED by kvcompress/synth/gen_synth_tables.py -- do not hand-edit.
\begin{wraptable}{r}{0.44\textwidth}
\vspace{-3.5\baselineskip}
\centering
\small
\caption{A passcode stated once, $57$ compression rounds before the question.}
\vspace{-1mm}
\label{tab:synthboundary}
\setlength{\tabcolsep}{4pt}
\begin{tabular}{lcc}
\toprule
Policy & Retr. & $\log p$ \\
\midrule
\method{} (ours) & $0.000$ & $-18.35$ \\
\vase{} & $0.344$ & $-3.88$ \\
\snapkv{} & $0.004$ & $-11.11$ \\
\rkv{} & $0.836$ & $-0.71$ \\
\triattn{} & $0.016$ & $-11.11$ \\
\bottomrule
\end{tabular}
\vspace{-1\baselineskip}
\end{wraptable}

The one case a signal-free policy cannot cover is a fact stated once,
never restated, and needed much later. We announce a passcode once, $57$
compression rounds before the question, let each policy retain as it
sees fit, and report two numbers (Table~\ref{tab:synthboundary}): the
fraction of traces in which the model reproduces the passcode (Retr.),
and the log-probability it assigns to the correct passcode at the
question, averaged over traces ($\log p$). A $\log p$ of $0$ means the
model would produce the passcode with certainty; a number like $-18$ means the passcode is effectively gone from the cache.
\method{} never reproduces it. \rkv{} finds it $84\%$ of the time, plausibly through its redundancy term, which favours keys unlike the rest of the cache, as a once-stated passcode is; \vase{} a third of the time; \snapkv{} and \triattn{} almost never. \citet{depthcache} prove that
random caches must lose at pointer-chasing when nothing is redundant.
Needle-finding is thus real selection skill, but it neither implies
nor follows from aggregate strength: \rkv{}, the best needle-finder, leads only one column of Table~\ref{tab:main}, while \triattn{}, the strongest baseline
there, recovers almost nothing here. On real traces the
case is rare, because the model keeps restating what it is still using.

\section{Efficiency Evaluation}
\label{sec:efficiency}

We measure efficiency in two settings: vLLM \citep{kwon2023efficient} with PagedAttention, following \triattn{}'s protocol, and batched decoding in HuggingFace Transformers without paging (\S\ref{sec:eval_protocol}), at the largest batch that fits on one GPU. The first compares \method{} with \triattn{} only: among the baselines only \triattn{} and \rkv{} have vLLM ports, and \citet{mao2026triattention} already show that \triattn{} is more efficient than \rkv{} at matched budget. The second compares all methods. Details of the two protocols are given in Appendix~\ref{sec:effdetail}.

% AUTO-GENERATED by kvcompress/eval/gen_paper_tables.py -- do not hand-edit.

\begin{table}[t]
\centering
\small
\setlength{\tabcolsep}{6pt}
\vspace{-3mm}
\caption{Serving throughput (output tok/s, and the multiple of full attention) under vLLM with PagedAttention on one H200 ($\budget{=}2048$, $1$k-token prompts, $32$k-token generations).}
\label{tab:vllm}
\begin{tabular}{lcccc}
\toprule
Method & Qwen3-4B & Phi-4-reasoning & Qwen3-14B & Qwen3-32B \\
\midrule
\dense{} & 1296\,($1.00\times$) & 780\,($1.00\times$) & 925\,($1.00\times$) & 346\,($1.00\times$) \\
\triattn{} & 1494\,($1.15\times$) & 1212\,($1.55\times$) & 1303\,($1.41\times$) & 700\,($2.02\times$) \\
\rowcolor{oursbg} \method{} (ours) & \best{2046\,($1.58\times$)} & \best{1737\,($2.23\times$)} & \best{1819\,($1.97\times$)} & \best{923\,($2.67\times$)} \\
\midrule
\rowcolor{oursbg} \emph{ours over \triattn{}} & $+37\%$ & $+43\%$ & $+40\%$ & $+32\%$ \\
\bottomrule
\end{tabular}
\end{table}

\paragraph{Under paged serving, \method{} is $32$--$43\%$ faster than \triattn{}.}
Both methods run on one H200 at $\budget{=}2048$ with $1$k-token prompts, $32$k-token generations and $128$ requests (at most $96$ concurrent on Qwen3-32B, whose weights leave a smaller KV pool; Appendix~\ref{sec:effdetail}). At $32$k tokens, full attention's cache fills the KV pool and caps how many requests run at once; compression raises that cap, so both methods beat full attention (\method{} by $1.6$--$2.7\times$ as shown in Table~\ref{tab:vllm}). With equal cache sizes and the same kernels, \method{}'s $32$--$43\%$ margin over \triattn{} comes from the scoring pass it skips (below); it also holds at capacity ($+41\%$, $+42\%$), at short generations and at half the load (Appendix~\ref{sec:effdetail}).

% At $32$k tokens per request the KV cache, not the compute, limits how
% many requests a GPU can hold at once, so a smaller cache means more
% requests in flight and higher throughput. \method{} serves
% $1.6$--$2.7\times$ the full-attention throughput across the four
% models, $32$--$43\%$ more than \triattn{} on the same kernels
% (Table~\ref{tab:vllm}).\footnote{These runs sit within $7\%$ of the capacity plateau: offering
% $512$ requests raises throughput by only $7\%$ on Qwen3-4B and moves it by under $1\%$ on Qwen3-14B; the margin over \triattn{} holds at capacity ($+41\%$ and
% $+42\%$).} The margin is not specific to this operating point: it holds at short
% generations, where compression itself does not yet pay, and at lighter
% loads.

\paragraph{At equal memory every evictor gains, and the scoring pass decides
the rest.} At the largest batch each method fits on one H200 ($\budget{=}3072$, $32$k generations; Figure~\ref{fig:isomem}, Appendix~\ref{sec:effdetail}), the compressed caches hold $109$--$200$ sequences against $20$--$28$ for full attention, the source of every evictor's $3$--$10\times$ speedup. \method{}, which computes no score, fits the largest batch and reaches $10.0\times$ and $8.8\times$ on Qwen3-4B and 14B ($28.8\times$ on Qwen3-4B at a smaller $\budget$ of $1024$). The \triattn{} row there is an unfused port, so the margin we claim is Table~\ref{tab:vllm}'s.

\paragraph{Why skipping the scoring pass is worth $32$--$43\%$ in serving.}
Timed alone, the pass is cheap. One per-layer eviction costs $0.30$\,ms under \method{}, which only compacts the cache, and $1.47$--$1.64$\,ms under \triattn{}, which scores it first; in a single stream the extra scoring amounts to a few percent of decoding time. Serving multiplies that small cost in two ways. First, the compressions pile up: with $128$ concurrent requests, each compressed every $64$ of its own tokens, vLLM compresses about two requests per decoding step ($61.6$k compressions over the $32$k run), and each compression happens at a synchronisation point between batched steps, so all $128$ requests wait while one is compressed. Second, content-dependent scoring is more expensive under paged serving than in plain batched decoding. Cache-statistic selectors need an extra pass over the paged KV state, and attention-weight selectors must recompute or explicitly expose attention statistics, because the fused kernels do not materialise them; \method{} needs neither and only compacts. On Qwen3-14B, \triattn{}'s $32$k run takes $910$\,s longer than \method{}'s, about $15$\,ms of whole-batch waiting per compression, against well under a millisecond for \method{}.

\section{Related Work}
\label{sec:related}

\paragraph{KV cache for long-context understanding.}
A major line of work on the KV cache targets long inputs and short outputs: one or many documents fill the cache at prefill and a short
answer follows. Three families reduce its memory. Quantization keeps
every token at lower precision \citep{liu2024kivi,hooper2024kvquant}.
Eviction discards tokens under a budget \citep{xu-etal-2025-refreshkv, park2025keydiff, xiao2025duoattention, li-etal-2026-real, an2026restkv}, scored by cumulative attention
\citep{zhang2023ho}, its persistence across steps
\citep{liu2023scissorhands}, or attention from a window of recent
queries \citep{li2024snapkv}, with budgets adapted per head or per layer
\citep{feng2025adakv,cai2024pyramidkv}, structural rules that keep the
attention sinks and a recent window \citep{xiao2024efficient}, or a
policy chosen per head \citep{ge2024model}. Query-aware selection
keeps every token and attends to a subset per query \citep{tang2024quest},
which saves compute but not memory. Randomness has also been studied in KV cache management.
At the serving-system level, \citet{wu2026randomization} randomly evict
unmarked leaf tokens from a prefix-sharing cache, making eviction more
robust to dynamic or adversarial query arrivals.
For long-context compression, \citet{kvec} sample for coverage of the
input, and \citet{protectiondominates} show, for long-context question
answering under a global cache cap, that once the prompt-boundary tokens
are guarded the choice of score is second-order and a random policy shares
in the recovery.

\paragraph{KV cache for long reasoning.}
Chains of thought invert the ratio: the prompt is a few hundred tokens
and the cache is filled during decoding. Sparse-attention selection has
been adapted to this regime \citep{gao2026sparse, yang2026less}, but it keeps the full cache in memory (\S\ref{sec:setting}); eviction is the only route that bounds it. Decode-time evictors generally estimate KV importance and use these scores to determine retention: \rkv{} uses a SnapKV-style score with a redundancy penalty so that restated content is not kept twice \citep{cai2025rkv}, \vase{} scores by value magnitude and fills the remaining budget stochastically \citep{chang2026valueaware}, following earlier score-guided randomized eviction \citep{chen-etal-2024-nacl}, and \triattn{} predicts each key's future attention from a calibrated query centre \citep{mao2026triattention}; LazyEviction estimates
future importance from recurring attention patterns
\citep{zhang-etal-2026-lazyeviction}. Learned evictors instead train a scorer for each base model, typically on domain-specific data: TrimKV \citep{bui2026cache} and DBTrimKV \citep{bui2026make} learn per-token retention scores that decay with age, and ForesightKV \citep{dong2026foresightkv} distills a future-attention oracle refined with GRPO. We compare only against training-free evictors and discuss how trained ones relate to our findings in Appendix~\ref{sec:discussion}. Concurrently with this work, Prefix Sliding \citep{muennighoff2026prefix} keeps only the prompt and a window of recent tokens, the recency+prompt policy of \S\ref{sec:mech_protect}, and also explores incorporating KV cache eviction into training, as \citet{kontonis2026memento} do. Evaluations of KV compression on reasoning report that eviction hurts chains of thought
more than long-context tasks and that random or recency baselines fall
far behind scored selection \citep{holdthought,semanticshbm}, and
\citet{chen-etal-2026-pitfalls} show that protocols differing in what
they protect make results across papers hard to compare. We run baselines at matched budget and show that when the prompt is kept, even evicting at random can be comparable to the strongest baseline while being much more efficient.

\section{Conclusion}
KV cache eviction for reasoning has been treated as a ranking problem:
estimate which cached tokens will matter and keep those. We find that
the ranking contributes almost nothing. A policy that keeps the prompt
and evicts uniformly at random within each head matches the strongest
baselines across four models and six tasks, and serves $32$--$43\%$
higher throughput in vLLM deployment. We explain this by showing that 1) the prompt is the fragile part of the cache, and once every method keeps it, most
of the gap between methods disappears; 2) the reasoning trace protects
itself through redundancy, in the text and across heads, so once the
prompt is safe a random draw keeps enough copies of what the model
still needs. What remains for
a selection signal is the rare fact stated once and never restated,
which reasoning seldom produces.

% \subsection*{AI use statement}
% % REQUIRED by the ICLR 2027 template; does not count toward the page limit.
% % TODO(user): review wording before submission.
% In this work, we used generative AI tools (LLM-based coding and writing
% assistants) to assist with drafting and editing the manuscript, implementing
% analysis and plotting code, and typesetting tables. We have not used
% generative AI tools to design the experiments, to select or register the
% hypotheses and thresholds, or to produce any reported measurement. All
% AI-assisted code and text were reviewed by the authors; reported statistics
% were recomputed from the released per-instance logs, and every registered
% prediction and its verdict is recorded in the paper independently of any
% AI-generated draft. We take responsibility for the final content of this
% work, including text, claims, and artifacts produced with the aid of
% generative AI.

\subsection*{Acknowledgements}
The authors would like to thank Zhiyi Shi and Haodong Wen for helpful discussion.

% \subsection*{Ethics statement}
% \subsection*{Reproducibility statement}

\bibliography{references}

@inproceedings{
li2024snapkv,
title={Snap{KV}: {LLM} Knows What You are Looking for Before Generation},
author={Yuhong Li and Yingbing Huang and Bowen Yang and Bharat Venkitesh and Acyr Locatelli and Hanchen Ye and Tianle Cai and Patrick Lewis and Deming Chen},
booktitle={The Thirty-eighth Annual Conference on Neural Information Processing Systems},
year={2024},
url={https://openreview.net/forum?id=poE54GOq2l}
}

@inproceedings{
cai2025rkv,
title={R-{KV}: Redundancy-aware {KV} Cache Compression for Reasoning Models},
author={Zefan Cai and Wen Xiao and Hanshi Sun and Cheng Luo and Yikai Zhang and Ke Wan and Yucheng Li and Yeyang Zhou and Li-Wen Chang and Jiuxiang Gu and Zhen Dong and Anima Anandkumar and Abedelkadir Asi and Junjie Hu},
booktitle={The Thirty-ninth Annual Conference on Neural Information Processing Systems},
year={2025},
url={https://openreview.net/forum?id=2jwAjomEDB}
}

@inproceedings{
chang2026valueaware,
title={Value-Aware Stochastic {KV} Cache Eviction for Reasoning Models},
author={Ting-Yun Chang and Harvey Yiyun Fu and Deqing Fu and Chenghao Yang and Jesse Thomason and Robin Jia},
booktitle={COLM 2026 Workshop on Efficient Reasoning},
year={2026},
url={https://openreview.net/forum?id=HysrOiWFIA}
}

@inproceedings{
mao2026triattention,
title={TriAttention: Efficient Long Reasoning with Trigonometric {KV} Compression},
author={Weian Mao and Xi Lin and Wei Huang and Yuxin Xie and Tianfu Fu and Bohan Zhuang and Song Han and Yukang Chen},
booktitle={Forty-third International Conference on Machine Learning},
year={2026},
url={https://openreview.net/forum?id=0tgzJK50Jz}
}

@inproceedings{
xiao2024efficient,
title={Efficient Streaming Language Models with Attention Sinks},
author={Guangxuan Xiao and Yuandong Tian and Beidi Chen and Song Han and Mike Lewis},
booktitle={The Twelfth International Conference on Learning Representations},
year={2024},
url={https://openreview.net/forum?id=NG7sS51zVF}
}

@inproceedings{
dao2024flashattention,
title={FlashAttention-2: Faster Attention with Better Parallelism and Work Partitioning},
author={Tri Dao},
booktitle={The Twelfth International Conference on Learning Representations},
year={2024},
url={https://openreview.net/forum?id=mZn2Xyh9Ec}
}

@inproceedings{kwon2023efficient,
  title={Efficient memory management for large language model serving with pagedattention},
  author={Kwon, Woosuk and Li, Zhuohan and Zhuang, Siyuan and Sheng, Ying and Zheng, Lianmin and Yu, Cody Hao and Gonzalez, Joseph and Zhang, Hao and Stoica, Ion},
  booktitle={Proceedings of the 29th symposium on operating systems principles},
  pages={611--626},
  year={2023}
}

@article{qwen3,
  title={Qwen3 Technical Report},
  author={An Yang and Anfeng Li and Baosong Yang and Beichen Zhang and Binyuan Hui and Bo Zheng and Bowen Yu and Chang Gao and Chengen Huang and Chenxu Lv and Chujie Zheng and Dayiheng Liu and Fan Zhou and Fei Huang and Feng Hu and Hao Ge and Haoran Wei and Huan Lin and Jialong Tang and Jian Yang and Jianhong Tu and Jianwei Zhang and Jianxin Yang and Jiaxi Yang and Jing Zhou and Jingren Zhou and Junyang Lin and Kai Dang and Keqin Bao and Kexin Yang and Le Yu and Lianghao Deng and Mei Li and Mingfeng Xue and Mingze Li and Pei Zhang and Peng Wang and Qin Zhu and Rui Men and Ruize Gao and Shixuan Liu and Shuang Luo and Tianhao Li and Tianyi Tang and Wenbiao Yin and Xingzhang Ren and Xinyu Wang and Xinyu Zhang and Xuancheng Ren and Yang Fan and Yang Su and Yichang Zhang and Yinger Zhang and Yu Wan and Yuqiong Liu and Zekun Wang and Zeyu Cui and Zhenru Zhang and Zhipeng Zhou and Zihan Qiu},
  journal={arXiv preprint arXiv:2505.09388},
  year={2025}
}

@article{phi4reasoning,
  title={Phi-4-reasoning Technical Report},
  author={Marah Abdin and Sahaj Agarwal and Ahmed Awadallah and Vidhisha Balachandran and Harkirat Behl and Lingjiao Chen and Gustavo de Rosa and Suriya Gunasekar and Mojan Javaheripi and Neel Joshi and Piero Kauffmann and Yash Lara and Caio C{\'e}sar Teodoro Mendes and Arindam Mitra and Besmira Nushi and Dimitris Papailiopoulos and Olli Saarikivi and Shital Shah and Vaishnavi Shrivastava and Vibhav Vineet and Yue Wu and Safoora Yousefi and Guoqing Zheng},
  journal={arXiv preprint arXiv:2504.21318},
  year={2025}
}

@article{deepseekr1,
  title={{DeepSeek-R1}: Incentivizing Reasoning Capability in {LLMs} via Reinforcement Learning},
  author={{DeepSeek-AI}},
  journal={arXiv preprint arXiv:2501.12948},
  year={2025}
}

@misc{openai2024o1,
    author       = {OpenAI},
    title        = {Learning to Reason with {LLM}s},
    year         = {2024},
    howpublished = {\url{https://openai.com/index/learning-to-reason-with-llms}},
  }

@inproceedings{
hendrycks2021measuring,
title={Measuring Mathematical Problem Solving With the {MATH} Dataset},
author={Dan Hendrycks and Collin Burns and Saurav Kadavath and Akul Arora and Steven Basart and Eric Tang and Dawn Song and Jacob Steinhardt},
booktitle={Thirty-fifth Conference on Neural Information Processing Systems Datasets and Benchmarks Track (Round 2)},
year={2021},
url={https://openreview.net/forum?id=7Bywt2mQsCe}
}

@inproceedings{
lightman2024lets,
title={Let's Verify Step by Step},
author={Hunter Lightman and Vineet Kosaraju and Yuri Burda and Harrison Edwards and Bowen Baker and Teddy Lee and Jan Leike and John Schulman and Ilya Sutskever and Karl Cobbe},
booktitle={The Twelfth International Conference on Learning Representations},
year={2024},
url={https://openreview.net/forum?id=v8L0pN6EOi}
}

@inproceedings{
rein2024gpqa,
title={{GPQA}: A Graduate-Level Google-Proof Q\&A Benchmark},
author={David Rein and Betty Li Hou and Asa Cooper Stickland and Jackson Petty and Richard Yuanzhe Pang and Julien Dirani and Julian Michael and Samuel R. Bowman},
booktitle={First Conference on Language Modeling},
year={2024},
url={https://openreview.net/forum?id=Ti67584b98}
}

@inproceedings{
balunovic2026matharena,
title={MathArena: Evaluating {LLM}s on Uncontaminated Math Competitions},
author={Mislav Balunovic and Jasper Dekoninck and Ivo Petrov and Nikola Jovanovi{\'c} and Martin Vechev},
booktitle={The Thirty-ninth Annual Conference on Neural Information Processing Systems Datasets and Benchmarks Track},
year={2026},
url={https://openreview.net/forum?id=y0zL9IZxZ7}
}

@inproceedings{
jain2025livecodebench,
title={LiveCodeBench: Holistic and Contamination Free Evaluation of Large Language Models for Code},
author={Naman Jain and King Han and Alex Gu and Wen-Ding Li and Fanjia Yan and Tianjun Zhang and Sida Wang and Armando Solar-Lezama and Koushik Sen and Ion Stoica},
booktitle={The Thirteenth International Conference on Learning Representations},
year={2025},
url={https://openreview.net/forum?id=chfJJYC3iL}
}

@inproceedings{
wu2026randomization,
title={Randomization Boosts {KV} Caching, Learning Balances Query Load: A Joint Perspective},
author={Fangzhou Wu and Sandeep Silwal and Qiuyi Zhang},
booktitle={The Fourteenth International Conference on Learning Representations},
year={2026},
url={https://openreview.net/forum?id=R7fv5NWfMm}
}

@article{kvec,
  title={Coverage-Driven {KV} Cache Eviction for Efficient and Improved Inference of {LLM}},
  author={Shuvendu Roy and Mengyao Zhai and Hossein Hajimirsadeghi and Golnoosh Samei},
  journal={arXiv preprint arXiv:2606.29563},
  year={2026}
}

@inproceedings{
wu2025retrieval,
title={Retrieval Head Mechanistically Explains Long-Context Factuality},
author={Wenhao Wu and Yizhong Wang and Guangxuan Xiao and Hao Peng and Yao Fu},
booktitle={The Thirteenth International Conference on Learning Representations},
year={2025},
url={https://openreview.net/forum?id=EytBpUGB1Z}
}

@inproceedings{
feng2025adakv,
title={Ada-{KV}: Optimizing {KV} Cache Eviction by Adaptive Budget Allocation for Efficient {LLM} Inference},
author={Yuan Feng and Junlin Lv and Yukun Cao and Xike Xie and S Kevin Zhou},
booktitle={The Thirty-ninth Annual Conference on Neural Information Processing Systems},
year={2025},
url={https://openreview.net/forum?id=tcisuhGsQZ}
}

@inproceedings{
xiao2025duoattention,
title={DuoAttention: Efficient Long-Context {LLM} Inference with Retrieval and Streaming Heads},
author={Guangxuan Xiao and Jiaming Tang and Jingwei Zuo and junxian guo and Shang Yang and Haotian Tang and Yao Fu and Song Han},
booktitle={The Thirteenth International Conference on Learning Representations},
year={2025},
url={https://openreview.net/forum?id=cFu7ze7xUm}
}

@article{depthcache,
  title={How Much Cache Does Reasoning Need? Depth-Cache Tradeoffs in {KV}-Compressed Transformers},
  author={Xiao Wang},
  journal={arXiv preprint arXiv:2604.17935},
  year={2026}
}

@article{protectiondominates,
  title={Protection Is (Nearly) All You Need: Structural Protection Dominates Scoring in Globally Capped {KV} Eviction},
  author={Gabriel Garcia},
  journal={arXiv preprint arXiv:2605.18053},
  year={2026}
}

@article{semanticshbm,
  title={Not All Thoughts Need {HBM}: Semantics-Aware Memory Hierarchy for {LLM} Reasoning},
  author={Aojie Yuan and Tianqi Shen and Dajun Zhang},
  journal={arXiv preprint arXiv:2605.09490},
  year={2026}
}

@article{holdthought,
  title={Hold Onto That Thought: Assessing {KV} Cache Compression on Reasoning},
  author={Minghui Liu and Aadi Palnitkar and Tahseen Rabbani and Hyunwoo Jae and Kyle Rui Sang and Dixi Yao and Shayan Shabihi and Fuheng Zhao and Tian Li and Ce Zhang and Furong Huang and Kunpeng Zhang},
  journal={arXiv preprint arXiv:2512.12008},
  year={2025}
}

@inproceedings{chen-etal-2026-pitfalls,
    title = "The Pitfalls of {KV} Cache Compression",
    author = "Chen, Alex  and
      Geh, Renato  and
      Grover, Aditya  and
      Van Den Broeck, Guy  and
      Israel, Daniel Mingyi",
    editor = "Liakata, Maria  and
      Moreira, Viviane P.  and
      Zhang, Jiajun  and
      Jurgens, David",
    booktitle = "Proceedings of the 64th Annual Meeting of the {A}ssociation for {C}omputational {L}inguistics (Volume 1: Long Papers)",
    month = jul,
    year = "2026",
    address = "San Diego, California, United States",
    publisher = "Association for Computational Linguistics",
    url = "https://aclanthology.org/2026.acl-long.1926/",
    doi = "10.18653/v1/2026.acl-long.1926",
    pages = "41530--41553",
    ISBN = "979-8-89176-390-6"
}

@inproceedings{
gao2026sparse,
title={Sparse Attention Adaptation for Long Reasoning},
author={Yizhao Gao and Shuming Guo and Shijie Cao and Yuqing Xia and Yu Cheng and Lei Wang and Lingxiao Ma and Yutao Sun and Tianzhu Ye and Li Dong and Hayden Kwok-Hay So and Yu Hua and Ting Cao and Fan Yang and Mao Yang},
booktitle={The Fourteenth International Conference on Learning Representations},
year={2026},
url={https://openreview.net/forum?id=c5BOcHM6J8}
}

@inproceedings{
zhang2023ho,
title={H2O: Heavy-Hitter Oracle for Efficient Generative Inference of Large Language Models},
author={Zhenyu Zhang and Ying Sheng and Tianyi Zhou and Tianlong Chen and Lianmin Zheng and Ruisi Cai and Zhao Song and Yuandong Tian and Christopher Re and Clark Barrett and Zhangyang Wang and Beidi Chen},
booktitle={Thirty-seventh Conference on Neural Information Processing Systems},
year={2023},
url={https://openreview.net/forum?id=RkRrPp7GKO}
}

@inproceedings{
tang2024quest,
title={{QUEST}: Query-Aware Sparsity for Efficient Long-Context {LLM} Inference},
author={Jiaming Tang and Yilong Zhao and Kan Zhu and Guangxuan Xiao and Baris Kasikci and Song Han},
booktitle={Forty-first International Conference on Machine Learning},
year={2024},
url={https://openreview.net/forum?id=KzACYw0MTV}
}

@inproceedings{
liu2023scissorhands,
title={Scissorhands: Exploiting the Persistence of Importance Hypothesis for {LLM} {KV} Cache Compression at Test Time},
author={Zichang Liu and Aditya Desai and Fangshuo Liao and Weitao Wang and Victor Xie and Zhaozhuo Xu and Anastasios Kyrillidis and Anshumali Shrivastava},
booktitle={Thirty-seventh Conference on Neural Information Processing Systems},
year={2023},
url={https://openreview.net/forum?id=JZfg6wGi6g}
}

@article{cai2024pyramidkv,
  title   = {{PyramidKV}: Dynamic {KV} Cache Compression based on Pyramidal Information Funneling},
  author  = {Cai, Zefan and Zhang, Yichi and Gao, Bofei and Liu, Yuliang and Liu, Tianyu and Lu, Keming and Xiong, Wayne and Dong, Yue and Chang, Baobao and Hu, Junjie and Xiao, Wen},
  journal = {arXiv preprint arXiv:2406.02069},
  year    = {2024}
}

@inproceedings{
ge2024model,
title={Model Tells You What to Discard: Adaptive {KV} Cache Compression for {LLM}s},
author={Suyu Ge and Yunan Zhang and Liyuan Liu and Minjia Zhang and Jiawei Han and Jianfeng Gao},
booktitle={The Twelfth International Conference on Learning Representations},
year={2024},
url={https://openreview.net/forum?id=uNrFpDPMyo}
}

@inproceedings{
liu2024kivi,
title={{KIVI}: A Tuning-Free Asymmetric 2bit Quantization for {KV} Cache},
author={Zirui Liu and Jiayi Yuan and Hongye Jin and Shaochen Zhong and Zhaozhuo Xu and Vladimir Braverman and Beidi Chen and Xia Hu},
booktitle={Forty-first International Conference on Machine Learning},
year={2024},
url={https://openreview.net/forum?id=L057s2Rq8O}
}

@inproceedings{
hooper2024kvquant,
title={{KVQ}uant: Towards 10 Million Context Length {LLM} Inference with {KV} Cache Quantization},
author={Coleman Richard Charles Hooper and Sehoon Kim and Hiva Mohammadzadeh and Michael W. Mahoney and Sophia Shao and Kurt Keutzer and Amir Gholami},
booktitle={The Thirty-eighth Annual Conference on Neural Information Processing Systems},
year={2024},
url={https://openreview.net/forum?id=0LXotew9Du}
}

@article{muennighoff2026prefix,
  title   = {Prefix Sliding for Efficient Test-Time Scaling},
  author  = {Muennighoff, Niklas and Wang, Zhengyang and Chen, Zeyi and Shi, Weijia and Hui, Binyuan and Yang, John and Jiang, Dapeng and Senghaas, Mika and Obeid, Fares and Hagemann, Johannes and Jaghouar, Sami and Schmidt, Ludwig and Liang, Percy and Wei, Jason and Ng, Andrew Y. and Zettlemoyer, Luke and Choi, Yejin and Lewis, Mike},
  journal = {arXiv preprint arXiv:2608.26070},
  year    = {2026}
}

@inproceedings{
yang2026less,
title={Less Is More: Fast and Accurate Reasoning with Cross-Head Unified Sparse Attention},
author={Lijie Yang and Zhihao Zhang and Arti Jain and Shijie Cao and Baihong Yuan and Yiwei Chen and Zhihao Jia and Ravi Netravali},
booktitle={Forty-third International Conference on Machine Learning},
year={2026},
url={https://openreview.net/forum?id=trSWJ99WzS}
}

@inproceedings{xu-etal-2025-refreshkv,
    title = "{R}efresh{KV}: Updating Small {KV} Cache During Long-form Generation",
    author = "Xu, Fangyuan  and
      Goyal, Tanya  and
      Choi, Eunsol",
    editor = "Che, Wanxiang  and
      Nabende, Joyce  and
      Shutova, Ekaterina  and
      Pilehvar, Mohammad Taher",
    booktitle = "Proceedings of the 63rd Annual Meeting of the Association for Computational Linguistics (Volume 1: Long Papers)",
    month = jul,
    year = "2025",
    address = "Vienna, Austria",
    publisher = "Association for Computational Linguistics",
    url = "https://aclanthology.org/2025.acl-long.1211/",
    doi = "10.18653/v1/2025.acl-long.1211",
    pages = "24878--24893",
    ISBN = "979-8-89176-251-0"
}

@inproceedings{
park2025keydiff,
title={KeyDiff: Key Similarity-Based {KV} Cache Eviction for Long-Context {LLM} Inference in Resource-Constrained Environments},
author={Junyoung Park and Dalton Jones and Matthew J Morse and Raghavv Goel and Mingu Lee and Christopher Lott},
booktitle={The Thirty-ninth Annual Conference on Neural Information Processing Systems},
year={2025},
url={https://openreview.net/forum?id=uBaFH7aQnC}
}

@inproceedings{
kontonis2026memento,
title={{MEMENTO}: Teaching {LLM}s to Manage Their  Context},
author={Vasilis Kontonis and Yuchen Zeng and Shivam Garg and Lingjiao Chen and Hao Tang and Ziyan Wang and Ahmed Hassan Awadallah and Eric Horvitz and John Langford and Dimitris Papailiopoulos},
booktitle={Third Conference on Language Modeling},
year={2026},
url={https://openreview.net/forum?id=YaYiQDVsi0}
}

@inproceedings{chen-etal-2024-nacl,
    title = "{NACL}: A General and Effective {KV} Cache Eviction Framework for {LLM} at Inference Time",
    author = "Chen, Yilong  and
      Wang, Guoxia  and
      Shang, Junyuan  and
      Cui, Shiyao  and
      Zhang, Zhenyu  and
      Liu, Tingwen  and
      Wang, Shuohuan  and
      Sun, Yu  and
      Yu, Dianhai  and
      Wu, Hua",
    editor = "Ku, Lun-Wei  and
      Martins, Andre  and
      Srikumar, Vivek",
    booktitle = "Proceedings of the 62nd Annual Meeting of the Association for Computational Linguistics (Volume 1: Long Papers)",
    month = aug,
    year = "2024",
    address = "Bangkok, Thailand",
    publisher = "Association for Computational Linguistics",
    url = "https://aclanthology.org/2024.acl-long.428/",
    doi = "10.18653/v1/2024.acl-long.428",
    pages = "7913--7926"
}

@inproceedings{han-etal-2024-lm,
    title = "{LM}-Infinite: Zero-Shot Extreme Length Generalization for Large Language Models",
    author = "Han, Chi  and
      Wang, Qifan  and
      Peng, Hao  and
      Xiong, Wenhan  and
      Chen, Yu  and
      Ji, Heng  and
      Wang, Sinong",
    editor = "Duh, Kevin  and
      Gomez, Helena  and
      Bethard, Steven",
    booktitle = "Proceedings of the 2024 Conference of the North American Chapter of the Association for Computational Linguistics: Human Language Technologies (Volume 1: Long Papers)",
    month = jun,
    year = "2024",
    address = "Mexico City, Mexico",
    publisher = "Association for Computational Linguistics",
    url = "https://aclanthology.org/2024.naacl-long.222/",
    doi = "10.18653/v1/2024.naacl-long.222",
    pages = "3991--4008"
}

@inproceedings{zhang-etal-2026-lazyeviction,
    title = "{L}azy{E}viction: Lagged {KV} Eviction with Attention Pattern Observation for Efficient Long Reasoning",
    author = "Zhang, Haoyue  and
      Zhang, Hualei  and
      Ma, Xiaosong  and
      Zhang, Jie  and
      Guo, Song",
    editor = "Liakata, Maria  and
      Moreira, Viviane P.  and
      Zhang, Jiajun  and
      Jurgens, David",
    booktitle = "Proceedings of the 64th Annual Meeting of the {A}ssociation for {C}omputational {L}inguistics (Volume 1: Long Papers)",
    month = jul,
    year = "2026",
    address = "San Diego, California, United States",
    publisher = "Association for Computational Linguistics",
    url = "https://aclanthology.org/2026.acl-long.1683/",
    doi = "10.18653/v1/2026.acl-long.1683",
    pages = "36335--36352",
    ISBN = "979-8-89176-390-6"
}

@inproceedings{li-etal-2026-real,
    title = "{REAL}: {RE}trieval-re{A}soning and Logic-constructed Attention Behaviors for Long-Context {KV} Cache Compression",
    author = "Li, Mengjie  and
      Feng, Yuan  and
      Xie, Xike  and
      Song, William J.",
    editor = "Liakata, Maria  and
      Moreira, Viviane P.  and
      Zhang, Jiajun  and
      Jurgens, David",
    booktitle = "Proceedings of the 64th Annual Meeting of the {A}ssociation for {C}omputational {L}inguistics (Volume 1: Long Papers)",
    month = jul,
    year = "2026",
    address = "San Diego, California, United States",
    publisher = "Association for Computational Linguistics",
    url = "https://aclanthology.org/2026.acl-long.1811/",
    doi = "10.18653/v1/2026.acl-long.1811",
    pages = "39035--39052",
    ISBN = "979-8-89176-390-6"
}

@inproceedings{
bui2026cache,
title={Cache What Lasts: Token Retention for Memory-Bounded {KV} Cache in {LLM}s},
author={Ngoc Bui and Shubham Sharma and Simran Lamba and Saumitra Mishra and Rex Ying},
booktitle={The Fourteenth International Conference on Learning Representations},
year={2026},
url={https://openreview.net/forum?id=qCaq3jGb0S}
}

@article{bui2026make,
  title={Make Each Token Count: Towards Improving Long-Context Performance with KV Cache Eviction},
  author={Bui, Ngoc and Nguyen, Hieu Trung and Cohan, Arman and Ying, Rex},
  journal={arXiv preprint arXiv:2605.09649},
  year={2026}
}

@inproceedings{
dong2026foresightkv,
title={Foresight{KV}: Optimizing {KV} Cache Eviction for Reasoning Models by Learning Long-Term Contribution},
author={Zican Dong and Peiyu Liu and Junyi Li and Zhipeng Chen and Han Peng and Shuo Wang and Xin Zhao},
booktitle={Forty-third International Conference on Machine Learning},
year={2026},
url={https://openreview.net/forum?id=znV8JHv8b8}
}

@inproceedings{
an2026restkv,
title={Re{ST}-{KV}: Robust {KV} Cache Eviction with Layer-wise Output Reconstruction and Spatial-Temporal Smoothing},
author={Yongqi An and Chang Lu and Kuan Zhu and Tao Yu and Chaoyang Zhao and Hong Wu and Ming Tang and Jinqiao Wang},
booktitle={The Fourteenth International Conference on Learning Representations},
year={2026},
url={https://openreview.net/forum?id=PhEHuo7oMm}
}
\bibliographystyle{tmlr}

\clearpage
\appendix
\section{Generality: additional models}
\label{sec:generality}

Table~\ref{tab:appendix_grid} repeats the main grid on Qwen3-14B, an
intermediate scale within the headline family. The picture from
Table~\ref{tab:main} replicates: \method{} beats \vase{} and \snapkv{} on
\mathfive{} and \gpqa{} (both significant). Three baseline cells are
significantly ahead at this scale: \triattn{} on \lcb{} ($+2.3$
points, $p{=}.007$), matching its code win on Qwen3-32B and the
prompt-length account of \S\ref{sec:main}, \triattn{} on \mathfive{}
($+2.1$ points, $p{=}.02$), and \vase{} on \aime{} ($+2.6$ points,
$p{=}.007$), consistent with its nominal \aime{} edge on Qwen3-32B.
Every cell here uses the same rollout count as the corresponding cell of
Table~\ref{tab:main} ($R{=}2$ on \mathfive{}, $4$ on \gpqa{} and \lcb{}, $16$ on
\hmmt{} and \aime{}) for every method.

% AUTO-GENERATED by kvcompress/eval/gen_paper_tables.py -- do not hand-edit.

\begin{table}[t]
\centering
\footnotesize
\setlength{\tabcolsep}{3.5pt}
\caption{Performance of Qwen3-14B with the same setting as in Table~\ref{tab:main}.}
\label{tab:appendix_grid}
\resizebox{0.9\textwidth}{!}{%
\begin{tabular}{lccccc}
\toprule
 & \mathfive{} & \gpqa{} & \aime{} & \hmmt{} & \lcb{}  \\
 & \scriptsize $K{=}1024$ & \scriptsize $K{=}2048$ & \scriptsize $K{=}4096$ & \scriptsize $K{=}4096$ & \scriptsize $K{=}3072$  \\
\midrule
\rowcolor{black!6}
\multicolumn{6}{l}{\textbf{Qwen3-14B}} \\
\dense{} & 0.951 & 0.645 & 0.706 & 0.557 & 0.856 \\
\noalign{\vskip -2pt}
\cmidrule(lr){1-6}
\snapkv{} & \sigbelow{0.812} & \sigbelow{0.506} & \sigbelow{0.460} & 0.467 & \sigbelow{0.622}  \\
\rkv{} & \sigbelow{0.816} & 0.600 & \sigbelow{0.571} & \sigbelow{0.427} & \sigbelow{0.788} \\
\vase{} & \sigbelow{0.852} & \sigbelow{0.543} & \best{0.668} & 0.495 & 0.813  \\
\triattn{} & \best{0.891} & 0.625 & 0.654 & 0.493 & \best{0.843}  \\
\rowcolor{oursbg}
\method{} (ours) & 0.870 &\best{0.628} & 0.642 & \best{0.505} & 0.820  \\
\bottomrule
\end{tabular}}
\end{table}

\section{The planted-fact probe: instrument, metric, and caveats}
\label{sec:synthapp}

\paragraph{Instrument.} Each instance is a short prefilled stub (chat
template plus think-opener, 23 tokens) followed by a fully scripted body fed
through the decode path token by token: real model-generated \mathfive{}
reasoning as filler (screened so it never contains the planted variable or
value), the fact inside a fixed 16-token box, a second fact 256 tokens later
(inert unless the cell uses it), and a terminal query whose value tokens are
the measurement points. Values are 4-digit numbers, distinct across
instances, and exactly four tokens under Qwen3's digit tokenizer. Because
every scripted token enters through the true decode path, eviction treats it
exactly as generated reasoning, and the model cannot restate the fact on
its own. Pinning gives the fact's span a $+\infty$ score in the
designated (layer, head) sites and $-\infty$ everywhere else (evicted at
first eligibility); background eviction is the standard per-head uniform
draw with no protection rule anywhere, and a per-event audit log records the
surviving set. Distances are eviction-event counts $E \in \{3,15,39,57\}$
with per-copy survival $(1024/1088)^{E}$. Four checks validate the instrument: (i) token-by-token forcing
matches one-shot prefill within kernel tolerance on every probe tested; (ii) the audit confirms the pinned
span survives in exactly the specified sites at every eviction event;
(iii) pinning is non-perturbative: keep-everywhere under eviction scores
at least as high as no-eviction ($+0.22$ nats); and (iv) the endpoints are
far apart (mean value logprob $-0.19$ nats with the fact kept everywhere
vs.\ $-13.59$ deleted everywhere; exact match $0.986$ vs.\ $0.000$).

\paragraph{Recall metric.} The graded recall of \S\ref{sec:mech_pool} is
\begin{equation}
R \;=\; \frac{\sum_i \bigl(\mathrm{LP}_i - \mathrm{LP}_i^{\mathrm{del}}\bigr)}
             {\sum_i \bigl(\mathrm{LP}_i^{\mathrm{kept}} - \mathrm{LP}_i^{\mathrm{del}}\bigr)},
\label{eq:recall}
\end{equation}
where $\mathrm{LP}_i$ is the log-probability the model assigns to trace $i$'s correct value under the condition being tested, and $\mathrm{LP}_i^{\mathrm{kept}}$ and $\mathrm{LP}_i^{\mathrm{del}}$ are the same quantity with the fact kept in every cache and deleted from every cache, both measured per needle. $R{=}1$ means the surviving copies are worth as much as never evicting the fact, and $R{=}0$ that they are worth nothing. Retrieval leads wherever it separates conditions; $R$ carries the comparison where retrieval floors, as it does on the two-needle probe.

\paragraph{Statistics.} Confidence intervals are instance-clustered
bootstrap percentile intervals with $4{,}000$ replicates; every condition
is scored on the same instances, so contrasts are paired. Re-running under fresh eviction randomness moves the reported values
by one to two points.

\paragraph{Caveats.} Exact retrieval has no dynamic range once a fact
survives in a single head (it floors near $0.03$), which is why
single-copy and two-fact contrasts are reported on $R$. The probe runs on Qwen3-4B; which heads act as carriers is a property
of the model, and no claim in the paper depends on their identity.
Replicating the probe on other models is left to future work.

\section{Matched protection in the remaining settings}
\label{sec:ppappendix}

% AUTO-GENERATED by kvcompress/eval/gen_paper_tables.py -- do not hand-edit.

\begin{table}[t]
\centering
\small
\setlength{\tabcolsep}{3pt}
\caption{Matched protection in the settings Table~\ref{tab:promptprotect} does not cover (\lcb{}: pass@1; others: accuracy; \aime{} pools 2025+2026). Small numbers give the gain from the rule in points, \gain{red} when at least $2$; bold marks the best protected method in each setting. \triattn{} and \method{} keep the prompt by construction and appear only in the protected column (values as in Table~\ref{tab:main}).}
\label{tab:promptprotect_appx}
\resizebox{\textwidth}{!}{%
\begin{tabular}{lcccccccccc}
\toprule
 & \multicolumn{6}{c}{Qwen3-4B} & \multicolumn{2}{c}{Phi-4-reasoning} & \multicolumn{2}{c}{Qwen3-32B} \\
\cmidrule(lr){2-7} \cmidrule(lr){8-9} \cmidrule(lr){10-11}
 & \multicolumn{2}{c}{\lcb{}} & \multicolumn{2}{c}{\aime{}} & \multicolumn{2}{c}{\hmmt{}} & \multicolumn{2}{c}{\lcb{}} & \multicolumn{2}{c}{\gpqa{}} \\
\cmidrule(lr){2-3} \cmidrule(lr){4-5} \cmidrule(lr){6-7} \cmidrule(lr){8-9} \cmidrule(lr){10-11}
Method & score alone & $+$ prompt & score alone & $+$ prompt & score alone & $+$ prompt & score alone & $+$ prompt & score alone & $+$ prompt \\
\midrule
\snapkv{} & 0.507 & 0.728\,\gain{+22.1} & 0.418 & 0.556\,\gain{+13.9} & 0.395 & 0.408\,\gainnull{+1.3} & 0.314 & 0.666\,\gain{+35.2} & 0.476 & 0.641\,\gain{+16.5} \\
\rkv{} & 0.712 & 0.698\,\gainnull{-1.4} & 0.494 & 0.499\,\gainnull{+0.5} & 0.371 & 0.372\,\gainnull{+0.1} & 0.621 & 0.630\,\gainnull{+0.9} & 0.638 & 0.639\,\gainnull{+0.1} \\
\vase{} & 0.700 & 0.732\,\gain{+3.2} & 0.596 & 0.575\,\gainnull{-2.1} & 0.421 & 0.411\,\gainnull{-1.0} & 0.373 & 0.646\,\gain{+27.3} & 0.597 & 0.626\,\gain{+2.9} \\
\midrule
\triattn{} & -- & \best{0.755} & -- & 0.592 & -- & 0.437 & -- & 0.652 & -- & \best{0.683} \\
\rowcolor{oursbg} \method{} & -- & 0.744 & -- & \best{0.610} & -- & \best{0.438} & -- & \best{0.667} & -- & \best{0.683} \\
\bottomrule
\end{tabular}}
\end{table}

Table~\ref{tab:promptprotect} gives every method the prompt-protection
rule on Qwen3-4B and Phi-4-reasoning for \mathfive{} and \gpqa{}.
Table~\ref{tab:promptprotect_appx} extends the same control to the
settings where Table~\ref{tab:main} shows the largest remaining
deficits: code on both models, competition math on Qwen3-4B, and
\gpqa{} at 32B. Two facts hold across all fifteen protected
model--task settings.

\paragraph{The payoff follows the retention deficit.} Every gain from
the rule is ordered by how much of the prompt the score was losing
(Appendix~\ref{sec:keeplog}). \snapkv{}, which retains the least, gains
$+35.2$, $+22.1$, $+16.5$, $+13.9$ and $+1.3$ points; \vase{} gains
$+27.3$, $+3.2$ and $+2.9$ where its retention falls short, and moves
by $-2.1$ and $-1.0$ on competition math, where it does not; \rkv{},
which already keeps most of the prompt, moves by $-1.4$ to $+0.9$
points, essentially zero, in all five settings. The smallest \snapkv{} gain is on \hmmt{}, where its unprotected
deficit is itself small at $32$k-token generations, so there is little
to close.

\paragraph{Protection closes most of the gap, not all of it.} Where the
deficit was prompt-driven the rule closes it completely: \snapkv{} and
\vase{} become statistical ties with \method{} on code for both models,
and \snapkv{} and \vase{} on \hmmt{}. Elsewhere a residual survives protection (paired
tests as in \S\ref{sec:eval_protocol}): all three protected baselines
remain $4$--$6$ points below \method{} on \gpqa{} at 32B, where
\triattn{} ties it exactly; \snapkv{}, \rkv{} and \vase{} remain $5.4$, $11.1$ and $3.5$ points
below on \aime{}, and \rkv{} $6.6$ below on \hmmt{}; and
\rkv{} remains $4.6$ and $3.7$ points below on code for both models even
though it never lost the prompt, a negative selection effect that
mirrors \triattn{}'s positive one on Qwen3-32B code. Since \triattn{}
and \method{} both keep the prompt by construction, every
\triattn{}-versus-\method{} cell of Table~\ref{tab:main} is already a
matched-protection comparison; the confound applies only to the other
baselines' deficits. The strongest reading of the 32B \gpqa{} setting is the
simplest: a uniform draw plus the rule beats all three protected scores there. We have not run the protected grid on Qwen3-14B.

\section{Keep-log measurements}
\label{sec:keeplog}

\begin{figure}[t]
\centering
\includegraphics[width=0.55\textwidth]{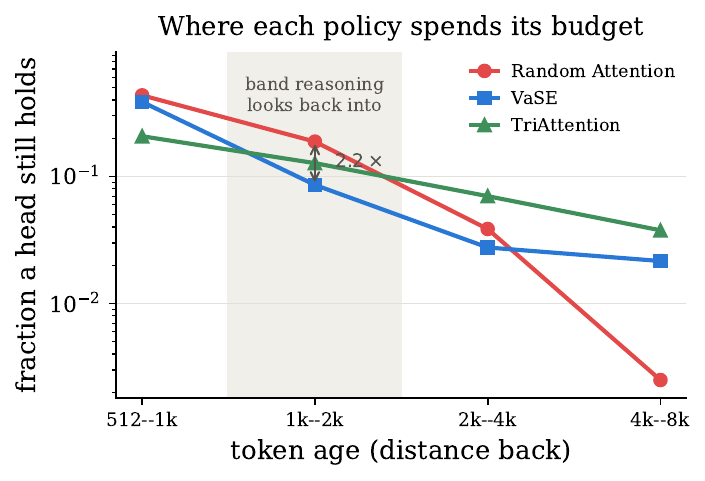}
\caption{Fraction of positions of a given age that a head still holds
(log scale; Qwen3-4B \mathfive{}, $\budget{=}1024$). \method{} decays
geometrically with age; \vase{} concentrates and freezes a tail of old
favourites; \triattn{} spends almost uniformly across ages, keeping
less of the recent middle than \method{} but several times more of the
very old tail.}
\label{fig:keeplog}
\end{figure}

The numbers in \S\ref{sec:mech_protect}--\S\ref{sec:mech_boundary} that
describe what a policy retains come from logging every eviction round of
real runs: nineteen policy--cell logs, $16$ traces each, of order $10^{4}$--$10^{5}$
rounds per log, recording for each round the keep-set of every
(layer, key-value head) pair together with the age of each retained position.
Three quantities are used. \emph{Slot coverage} is the fraction of
candidate positions retained by at least one head immediately after a round;
it is $0.999$--$1.000$ for \method{} and \vase{}, $0.993$ for \triattn{},
and $0.938$ for the shared-draw control, which by construction has no
cross-head diversity. \emph{Prompt survival} is the fraction of prompt
positions still held, by any head and by a given head: $0.994$--$0.999$ and
the same per head for \method{}, against $0.55$--$0.91$ (union) and
$0.26$--$0.67$ (per head) for \rkv{}, $0.56$--$0.70$ and $0.20$--$0.29$
for \vase{}, and $0.32$--$0.42$ and $0.11$--$0.22$ for \snapkv{}, the
lowest on every one of the four model--task settings. \emph{Survival by age} is the fraction of positions in an
age band a head still holds (Figure~\ref{fig:keeplog}); at $1$--$2$k it
is $0.188$ for \method{},
$0.161$ for the shared draw, $0.145$ for \snapkv{}, $0.105$ for \rkv{} and
$0.086$ for \vase{}, and the cross-head union in that band is $0.776$ for
\method{} against $0.368$ for \vase{} and $0.161$ for the shared draw,
whose union cannot exceed its per-head survival. \triattn{} holds $0.127$ per head in that band
with a cross-head union of only $0.199$: it pins the prompt in full, and
its heads keep nearly the same positions, the least cross-head
diversity of any per-head policy measured here.

\paragraph{Implicit age bias.} Although no score is computed, the policy
is not age-blind. A position that survives one eviction faces a fresh
draw at the next, so its probability of still being cached after $n$
evictions is $\big(\tfrac{\budget-\promptlen}{\budget+\residual-\promptlen}\big)^{n}$,
about $0.94^{n}$ at $\budget{=}1024$ and $\residual{=}64$. \method{}
is therefore a soft recency window: recent positions are almost always
present, and old ones survive as a thin tail that differs from head to
head (Figure~\ref{fig:keeplog}). \S\ref{sec:mech_protect} separates
the two ingredients, a hard recency window with the prompt kept and the
tail that the scatter adds.

\paragraph{Accuracy of the shared draw.} The most direct task-level test
of cross-head diversity is to remove it: the shared-draw control keeps
the prompt and draws one random keep-set that every head uses, so its
cross-head union equals its per-head survival ($0.161$ in the $1$--$2$k
band against $0.776$ for \method{}). On Qwen3-4B \mathfive{} it scores
$0.871$ at $\budget{=}1024$ and $0.788$ at $\budget{=}512$, against
$0.874$ and $0.789$ for \method{}. Removing cross-head diversity
therefore costs nothing on real traces at these budgets: the text-level
redundancy already keeps a restated copy of what the model still needs,
so the second level is not load-bearing there. It is load-bearing where
the first level is absent (a fact stated once and never restated), which
is the regime of the planted-fact probe (\S\ref{sec:mech_pool}) and of
the boundary in \S\ref{sec:mech_boundary}. The two redundancies are
substitutes, and \method{} preserves both.

\section{Generation lengths and run-to-run variability}
\label{sec:lensvar}

Table~\ref{tab:genlens} reports the mean number of generated tokens for
every cell of Tables~\ref{tab:main} and~\ref{tab:appendix_grid}, and
Table~\ref{tab:runvar} the standard deviation of per-run accuracy
across each cell's independently sampled runs. Two facts are worth
noting. First, the budgets of \S\ref{sec:eval_protocol} track the
full-attention lengths: \mathfive{} traces are the shortest and
competition-math traces the longest on every model, and eviction
generally lengthens generation relative to full attention, most for the
weakest selectors. Averaged over the five tasks, \method{} is the
shortest-generating evictor on Qwen3-4B and Qwen3-14B and within about
$5\%$ of the shortest on the other two models, so its accuracy parity
is not bought with longer generations. Second, the variability matches the significance
treatment in the main text: run-to-run standard deviation is at or
under about one point on \mathfive{}, one to three points on \gpqa{}, up to two on \lcb{}, and two to five points on competition math, which is why
every claim is decided by the paired, problem-clustered tests of
\S\ref{sec:eval_protocol} rather than by raw cell differences.
\mathfive{} cells have $R{=}2$, so their standard deviation is a
two-sample estimate.

% AUTO-GENERATED by kvcompress/analysis/build_lens_variance_tables.py -- do not hand-edit.

\begin{table}[t]
\centering
\footnotesize
\setlength{\tabcolsep}{3.5pt}
\caption{Mean generated tokens (thousands) per cell of Tables~\ref{tab:main} and~\ref{tab:appendix_grid}, measured over every run of the cell; Avg = unweighted mean over the five tasks.}
\label{tab:genlens}
\resizebox{0.9\textwidth}{!}{%
\begin{tabular}{lcccccc}
\toprule
 & \mathfive{} & \gpqa{} & \aime{} & \hmmt{} & \lcb{} & Avg \\
\midrule
\rowcolor{black!7}
\multicolumn{7}{l}{\textbf{Qwen3-4B}} \\
\dense{} & 5.3 & 8.6 & 17.0 & 18.4 & 11.1 & 12.1 \\
\noalign{\vskip -2pt}
\cmidrule(lr){1-7}
\snapkv{} & 14.2 & 12.0 & 20.1 & 22.7 & 15.8 & 17.0 \\
\rkv{} & 9.1 & 12.6 & 21.5 & 23.3 & 14.6 & 16.2 \\
\vase{} & 8.3 & 11.6 & 18.9 & 21.3 & 14.3 & 14.9 \\
\triattn{} & 8.0 & 11.3 & 19.4 & 21.2 & 13.6 & 14.7 \\
\rowcolor{oursbg}
\method{} (ours) & 7.1 & 9.9 & 18.6 & 20.0 & 14.1 & 13.9 \\
\midrule
\rowcolor{black!7}
\multicolumn{7}{l}{\textbf{Phi-4-reasoning}} \\
\dense{} & 2.7 & 7.5 & 13.0 & 16.5 & 9.9 & 9.9 \\
\noalign{\vskip -2pt}
\cmidrule(lr){1-7}
\snapkv{} & 2.9 & 8.6 & 13.3 & 17.0 & 10.5 & 10.4 \\
\rkv{} & 3.1 & 10.7 & 13.9 & 18.1 & 11.3 & 11.4 \\
\vase{} & 2.8 & 9.0 & 13.4 & 17.0 & 10.7 & 10.6 \\
\triattn{} & 3.3 & 9.4 & 13.9 & 17.3 & 10.5 & 10.9 \\
\rowcolor{oursbg}
\method{} (ours) & 3.4 & 9.4 & 13.4 & 17.3 & 10.5 & 10.8 \\
\midrule
\rowcolor{black!7}
\multicolumn{7}{l}{\textbf{Qwen3-32B}} \\
\dense{} & 4.6 & 7.0 & 14.8 & 17.4 & 10.3 & 10.8 \\
\noalign{\vskip -2pt}
\cmidrule(lr){1-7}
\snapkv{} & 7.5 & 10.1 & 17.2 & 20.4 & 15.0 & 14.0 \\
\rkv{} & 7.9 & 10.3 & 18.6 & 22.1 & 14.5 & 14.7 \\
\vase{} & 7.0 & 9.7 & 16.8 & 20.0 & 15.3 & 13.7 \\
\triattn{} & 7.2 & 8.3 & 17.0 & 20.0 & 12.8 & 13.0 \\
\rowcolor{oursbg}
\method{} (ours) & 6.1 & 8.0 & 16.7 & 19.2 & 18.4 & 13.7 \\
\midrule
\rowcolor{black!7}
\multicolumn{7}{l}{\textbf{Qwen3-14B}} \\
\dense{} & 4.8 & 7.6 & 15.7 & 17.9 & 9.8 & 11.2 \\
\noalign{\vskip -2pt}
\cmidrule(lr){1-7}
\snapkv{} & 6.9 & 10.4 & 18.3 & 20.1 & 14.2 & 14.0 \\
\rkv{} & 8.7 & 11.1 & 19.8 & 22.7 & 13.3 & 15.1 \\
\vase{} & 8.0 & 11.1 & 17.7 & 21.0 & 11.7 & 13.9 \\
\triattn{} & 7.1 & 9.4 & 17.8 & 20.5 & 11.6 & 13.3 \\
\rowcolor{oursbg}
\method{} (ours) & 6.9 & 8.8 & 17.6 & 20.1 & 12.3 & 13.1 \\
\bottomrule
\end{tabular}}
\end{table}

% AUTO-GENERATED by kvcompress/analysis/build_lens_variance_tables.py -- do not hand-edit.

\begin{table}[t]
\centering
\footnotesize
\setlength{\tabcolsep}{3.5pt}
\caption{Run-to-run variability: standard deviation of per-run accuracy (points) across each cell's independent sampled runs, including \lcb{}, whose runs are graded individually by test execution.}
\label{tab:runvar}
\resizebox{0.9\textwidth}{!}{%
\begin{tabular}{lccccc}
\toprule
 & \mathfive{} & \gpqa{} & \aime{} & \hmmt{} & \lcb{} \\
\midrule
\rowcolor{black!7}
\multicolumn{6}{l}{\textbf{Qwen3-4B}} \\
\dense{} & 0.1 & 1.1 & 4.3 & 4.8 & 0.5 \\
\noalign{\vskip -2pt}
\cmidrule(lr){1-6}
\snapkv{} & 1.3 & 2.9 & 2.9 & 3.5 & 2.0 \\
\rkv{} & 0.0 & 2.0 & 1.9 & 3.2 & 1.4 \\
\vase{} & 0.1 & 1.9 & 2.5 & 3.3 & 1.8 \\
\triattn{} & 0.2 & 2.5 & 5.2 & 3.9 & 0.8 \\
\rowcolor{oursbg}
\method{} (ours) & 0.0 & 1.4 & 3.7 & 2.7 & 1.5 \\
\midrule
\rowcolor{black!7}
\multicolumn{6}{l}{\textbf{Phi-4-reasoning}} \\
\dense{} & 0.4 & 2.6 & 3.6 & 3.1 & 1.4 \\
\noalign{\vskip -2pt}
\cmidrule(lr){1-6}
\snapkv{} & 0.8 & 1.5 & 3.2 & 3.6 & 0.5 \\
\rkv{} & 0.1 & 3.3 & 3.6 & 5.5 & 1.3 \\
\vase{} & 0.7 & 2.4 & 3.1 & 3.7 & 0.9 \\
\triattn{} & 1.3 & 1.4 & 5.1 & 4.6 & 2.0 \\
\rowcolor{oursbg}
\method{} (ours) & 1.0 & 2.3 & 3.3 & 3.5 & 1.4 \\
\midrule
\rowcolor{black!7}
\multicolumn{6}{l}{\textbf{Qwen3-32B}} \\
\dense{} & 0.6 & 1.2 & 3.3 & 4.1 & 0.8 \\
\noalign{\vskip -2pt}
\cmidrule(lr){1-6}
\snapkv{} & 0.6 & 1.8 & 3.2 & 4.0 & 1.7 \\
\rkv{} & 0.1 & 2.8 & 3.0 & 2.3 & 0.3 \\
\vase{} & 0.2 & 1.7 & 3.7 & 4.4 & 0.7 \\
\triattn{} & 0.7 & 1.2 & 3.7 & 3.9 & 1.0 \\
\rowcolor{oursbg}
\method{} (ours) & 0.3 & 1.1 & 3.8 & 4.3 & 1.6 \\
\midrule
\rowcolor{black!7}
\multicolumn{6}{l}{\textbf{Qwen3-14B}} \\
\dense{} & 0.1 & 2.2 & 2.9 & 3.2 & 1.2 \\
\noalign{\vskip -2pt}
\cmidrule(lr){1-6}
\snapkv{} & 0.4 & 2.3 & 3.9 & 3.2 & 1.4 \\
\rkv{} & 0.0 & 1.8 & 3.4 & 3.6 & 0.6 \\
\vase{} & 0.7 & 1.5 & 3.8 & 3.1 & 2.0 \\
\triattn{} & 0.7 & 1.6 & 4.4 & 3.3 & 0.8 \\
\rowcolor{oursbg}
\method{} (ours) & 0.5 & 2.6 & 3.7 & 3.4 & 1.3 \\
\bottomrule
\end{tabular}}
\end{table}

\section{Engine details}
\label{sec:engine}

We implement per-KV-head physical eviction on the \vase{} engine: every
$\residual{=}64$ decode steps, once the cache exceeds
$\budget{+}\residual$, each KV head scores its candidate slots (all cached
positions except the $\residual$-token recent buffer), keeps
$\operatorname{top-}\budget$, and the KV tensors are compacted with a gather.
Keys are stored post-RoPE with a cumulative-length counter; keep-sets are
sorted chronologically before compaction for every method; grouped-query
models evict per KV head (query groups share their head's keep-set). Eviction
is monotonic. Sampling and prompting follow the reference repository defaults
for every method; the model decodes until end-of-sequence, up to a uniform $32$k-token limit.

\paragraph{Baseline configurations.} Every baseline runs in this engine
at the same budget, trigger and buffer. \vase{} uses
$n_{\mathrm{large}}=\budget/4$, the setting in its released run scripts
(a fixed $n_{\mathrm{large}}{=}256$ at larger budgets degrades it toward
a recency policy). \rkv{} uses $\lambda{=}0.5$, the setting behind the
R-KV rows in \citet{chang2026valueaware}, which our cells reproduce to within $2.3$
points on \mathfive{} and $0.8$ on \gpqa{}; the recommended
$\lambda{=}0.1$ scores $7$--$9$ points lower on Qwen3-4B in both our
port and a line-by-line re-implementation of the official repository.
\triattn{} is ported verbatim from the official implementation in its
stronger per-head variant, with per-model calibration statistics that
reproduce the reference ranking (mean reciprocal rank $0.99$); its
released harness evaluates reasoning models with the chat template
disabled, which is why we do not compare against the paper's reported
numbers. Our full-attention path reproduces the full-attention
accuracies reported by \citet{chang2026valueaware} on Qwen3-4B to within $0.5$ points
on all four shared tasks. All cells are graded at the $32$k-token limit
by the final boxed answer (\lcb{}: by test execution), and no partially
completed cell enters any table.

\paragraph{Hardware.} Accuracy generation ran on a mixed fleet of
NVIDIA H200 nodes. Every efficiency measurement in
\S\ref{sec:efficiency} and Appendix~\ref{sec:effdetail} is H200-only,
with a single job per GPU and nothing else on the node.

\paragraph{Per-task budgets.} The main grid sets \mathfive{}
$\budget{=}1024$, \gpqa{} $\budget{=}2048$, AIME/\hmmt{}
$\budget{=}4096$, and \lcb{} $\budget{=}3072$, ${\sim}3\times$ compression of its ${\sim}10$k-token full-attention traces.

\section{Efficiency: protocols and additional measurements}
\label{sec:effdetail}

\paragraph{The vLLM setting (Table~\ref{tab:vllm}).}
Every run is bf16 on one H200, vLLM v0.19.0 with PagedAttention (CUDA graphs and prefix caching disabled), budget $2048$ and $1$k-token prompts, inside
the compression integration released with \triattn{}, which we run
unmodified except for the selector: \method{} is added as a scoring
function inside it, so both methods share the same attention kernels,
paging, scheduler and compression trigger and differ only in which positions
are kept, with a request compressed every $64$ generated tokens.
Compression is active throughout every compressed run ($55{,}296$
eviction events at the $8$k point and $61{,}568$ at the $32$k point,
identical for the two methods, with no skipped round and no prefix
mismatch; $59{,}008$ at the $32$k point for Phi-4-reasoning), and the same integration reproduces our accuracy at the same
cadence ($0.864$ on Qwen3-4B \mathfive{} at $\budget{=}1024$). The
compressed cells are single runs; repeating the benchmark with fresh
seeds, two to three per cell in earlier rounds at these load points,
moved every arm by at most $1.1\%$ (Qwen3-14B \method{} by $0.05\%$),
and the full-attention cells, which the plugin never touches, are means
of two runs with three repetitions agreeing within $0.4\%$; run-to-run
variability is therefore about one percent, two orders below the
margins we report.
Qwen3-32B runs
the same protocol; its $64$\,GB of weights shrink
the free KV pool, which lowers the preemption-safe cap (below) and, at
$32$k generations, where a full-length full-attention sequence carries
over $8$\,GB of KV, drops full-attention throughput to $346$ tok/s.
Phi-4-reasoning also runs the same protocol, with $31.5$k-token
generations, the most its $32$k context allows after the $1$k-token
prompt.
The $128$-request point is close to the paged capacity plateau, the
regime in which \triattn{} report their headline multiple: offering
$512$ requests on Qwen3-4B with \method{} at its preemption-safe
ceiling of $224$ resident sequences (verified, zero preemptions) raises
steady-state throughput from $2046$ to $2188$ tok/s, $+7\%$; the
decoding step there fits $9.5$\,ms $+ 0.415$\,ms per resident
sequence, so throughput is already $85\%$ batch-proportional at $128$
and cannot rise by more than $18\%$ at any batch. On Qwen3-14B the
same probe leaves throughput within $1\%$ of the $128$-request row
($1817$ vs.\ $1819$ tok/s). With both methods at the ceiling the margin
holds: $2117$ vs.\ $1501$ tok/s on Qwen3-4B ($+41\%$) and $1817$ vs.\
$1276$ on Qwen3-14B ($+42\%$), against $+37\%$ and $+40\%$ at $128$
requests. On Qwen3-32B the $128$-request run already operates at its
$96$-sequence ceiling, so no separate probe is needed.

\paragraph{Short generations, where compression does not pay.} A second
operating point offers $512$ requests of $1$k in, $8$k out, with the
compressed runs at their preemption-safe caps ($224$ on Qwen3-4B and 14B,
$96$ on Qwen3-32B) and full attention uncapped. There the arithmetic,
not the cache, bounds throughput, and a compressed cache buys nothing:
\method{} serves $0.52\times$, $0.70\times$ and $0.96\times$ the
full-attention throughput on Qwen3-4B, 14B and 32B, and $0.76\times$ on
Phi-4-reasoning; \triattn{} $0.38\times$, $0.49\times$, $0.70\times$
and $0.57\times$. The deficit shrinks with
model size because per-token KV grows with the model while the weights
shrink the cache pool, so the same workload becomes memory-bound as the
model grows, reaching near parity on Qwen3-32B. Phi-4-reasoning, with
the parameter count of Qwen3-14B but a quarter more KV per token ($40$
layers of $10$ KV heads against $40$ of $8$), sits between Qwen3-14B and
Qwen3-32B, as this account predicts. The margin between the
two compressed methods is unchanged there: \method{} leads \triattn{} by
$+39\%$, $+42\%$ and $+36\%$ on the Qwen3 models and $+35\%$ on
Phi-4-reasoning. The $64$-request and single-request
measurements keep the same request shape with the offered load reduced
to $64$ and $1$. At the $64$-request load \method{} leads \triattn{} by
$+35\%$ on Qwen3-4B and $+30\%$ on Qwen3-14B; at a single request the
two are within about one percent of each other ($115.8$ vs.\ $117.0$\,s
on Qwen3-4B, $125.8$ vs.\ $126.9$\,s on Qwen3-14B), consistent with
the sub-$2$\,ms calls of Table~\ref{tab:evictcost}: the serving
margin is the barrier-multiplied cost of reading paged cache state, not
a kernel-time gap, and an integration that scored asynchronously, off
the barrier, could shrink it; none is released, and \method{} needs
none. Two pitfalls in the released tooling silently make it measure the wrong thing
and are worth recording: vLLM's own throughput benchmark ignores the
requested output length (its default $128$-token outputs never reach
the compression threshold, so it times full-attention decoding), and
the integration's deduplication guard can disable all later compaction
after one benign under-budget round. We corrected both with bookkeeping
changes that leave selection and kernel semantics untouched; every
reported run is verified by its applied-event counters.

\triattn{} report ${\sim}2.5\times$ over full attention from a capacity
measurement taken at a different budget and decode length, on a different GPU; our
$1.2$--$2.0\times$ for their method sits below it, and the comparison we
draw is between the two methods we measured rather than against their published
figure.

The paged table runs at $\budget{=}2048$ and the equal-memory table
(Table~\ref{tab:isomem}) at $3072$; the $32$k
point rerun at
$\budget{=}3072$ preserves the ordering (\method{}
$2011$ and $1696$ tok/s against \triattn{}'s $1437$ and $1223$ on Qwen3-4B
and 14B, $+40\%$ and $+39\%$), so the budget difference between the two
tables drives neither comparison.

\paragraph{The concurrency cap.} When the KV pool is oversubscribed, vLLM V1
preempts requests silently, and the integration's compression state does not
survive preemption: on resume it compacts inside block IDs the request no
longer owns, and the scheduler's consistency check then kills the
engine. Capacity measurements therefore need a cap that keeps the pool
undersubscribed; ours is $224$ requests on Qwen3-4B and 14B and $96$ on
Qwen3-32B at the $8$k capacity point, whose $512$ offered requests keep
the pool under sustained admission pressure. The $32$k point offers $128$
requests in total and runs preemption-free at that concurrency on
Qwen3-4B and 14B; on Qwen3-32B it is capped at $96$. An uncapped
compressed run under sustained oversubscription is
not possible in this integration. Both compressed methods run under the
identical cap, so the comparison between them is unaffected. Full attention
tolerates preemption (vLLM recomputes the evicted request) and runs
uncapped, which favours it on the two smaller models; on Qwen3-32B the
$512$ offered requests oversubscribe its pool, so part of its deficit there
may be preemption overhead rather than capacity alone, and the comparison
we rest on at that scale is the cap-matched one against \triattn{}. The
$32$k full-attention baselines are oversubscribed in the same way on every model,
so the full-attention multiples of Table~\ref{tab:vllm} bundle capacity with
full attention's preemption overhead; the margins over \triattn{}, measured under
identical load, are unaffected.

% AUTO-GENERATED by kvcompress/eval/gen_paper_tables.py -- do not hand-edit.

\begin{table}[t]
\centering
\small
\caption{Cost of one eviction round (scoring $+$ compaction), measured with CUDA events on an otherwise idle H200: $\budget{=}1024$, $4096$ decode steps, single stream. \method{} performs no scoring, so its round time is the compaction floor every evictor pays; the excess over it is the price of the selection signal. The ordering, and the per-call costs to within $12\%$, are unchanged across a $3.5\times$ change in model size.}
\label{tab:evictcost}
\begin{tabular}{lcccc}
\toprule
 & \multicolumn{2}{c}{Qwen3-4B} & \multicolumn{2}{c}{Qwen3-14B} \\
\cmidrule(lr){2-3}\cmidrule(lr){4-5}
Method & ms\,/\,round & \% of decode & ms\,/\,round & \% of decode \\
\midrule
\rowcolor{oursbg} \method{} (ours) & 0.30 & 0.57\% & 0.29 & 0.53\% \\
\snapkv{} & 0.37 & 0.68\% & 0.39 & 0.68\% \\
\rkv{} & 0.58 & 1.06\% & 0.61 & 1.07\% \\
\vase{} & 0.74 & 1.35\% & 0.77 & 1.37\% \\
\triattn{} & 1.47 & 2.49\% & 1.64 & 2.67\% \\
\bottomrule
\end{tabular}
\end{table}

\paragraph{Eviction-round timing (Table~\ref{tab:evictcost}).} The full
round, scoring and compaction, is timed with CUDA events on an
otherwise idle node, single stream, $\budget{=}1024$, $4096$ decode steps,
$1872$ ($4$B) and $2080$ ($14$B) eviction calls.

\begin{figure}[t]
\centering
\includegraphics[width=\textwidth]{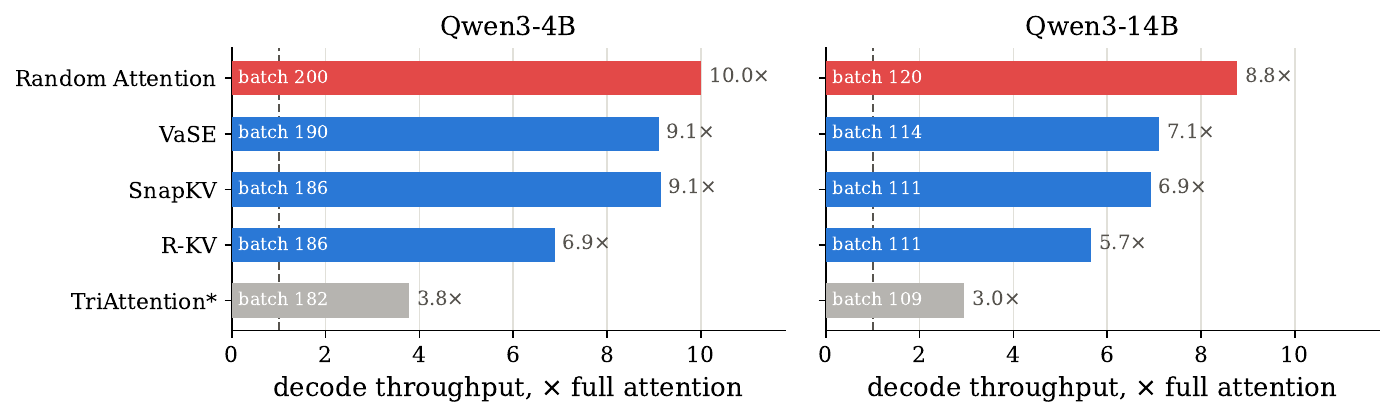}
\caption{Equal-memory serving: decode throughput relative to full
attention at each method's largest batch on one H200
($\budget{=}3072$, $32$k generations). $^*$\triattn{} here is an
unfused re-implementation of its scorer, far slower than the vLLM
version.}
\label{fig:isomem}
\end{figure}

% AUTO-GENERATED by kvcompress/eval/gen_paper_tables.py -- do not hand-edit.

\begin{table}[t]
\centering
\small
\setlength{\tabcolsep}{5pt}
\caption{Serving throughput when each method runs at the largest batch that fits one $143$\,GB H200, at $\budget{=}3072$ with $32$k generations. The small cache is what buys the batch, so every evictor collects most of the win; the ordering among them follows the cost of their scoring pass. \textbf{The \triattn{} row is an unfused re-implementation of its scorer, far slower than the authors' kernels, on which the gap to \method{} on these two models is $1.4\times$, not $2.7$--$3.0\times$ (Table~\ref{tab:vllm}).} Every other row runs on one shared code path.}
\label{tab:isomem}
\begin{tabular}{lcccccc}
\toprule
 & \multicolumn{3}{c}{Qwen3-4B} & \multicolumn{3}{c}{Qwen3-14B} \\
\cmidrule(lr){2-4} \cmidrule(lr){5-7}
Method & batch & tok/s & $\times$\,full & batch & tok/s & $\times$\,full \\
\midrule
\dense{} (no eviction) & 28 & 178 & 1.00 & 20 & 164 & 1.00 \\
\snapkv{} & 186 & 1624 & 9.14 & 111 & 1133 & 6.92 \\
\rkv{} & 186 & 1223 & 6.88 & 111 & 925 & 5.65 \\
\vase{} & 190 & 1617 & 9.10 & 114 & 1162 & 7.10 \\
\triattn{} & 182 & 670 & 3.77 & 109 & 483 & 2.95 \\
\rowcolor{oursbg} \method{} (ours) & 200 & \best{1779} & \best{10.01} & 120 & \best{1436} & \best{8.78} \\
\bottomrule
\end{tabular}
\end{table}

\paragraph{The equal-memory comparison (Table~\ref{tab:isomem}, i.e.\
Figure~\ref{fig:isomem} with exact throughput).} Every method
runs on the same engine and code path, so the comparison among them isolates
the scoring pass; only the batch differs, found per method by bisecting for
the largest that fits a $143$\,GB H200. Full attention needs its own search at each
decode length (at $14$B and $32$k, batches of $40$, $38$ and $34$ all
run out of memory and the largest feasible is $20$, at $129.2$\,GB peak). One caveat applies to \triattn{} and to no other selector: its row
runs our PyTorch port, and their paper states that no optimised
kernel exists yet, so the row reflects an unfused implementation rather than
the speed \triattn{} can attain, which is why the method-level comparison
in \S\ref{sec:efficiency} is the vLLM one, run on the kernels they release. No throughput number in this paper comes from the batched
accuracy runs, where several workers share a GPU.

% AUTO-GENERATED by kvcompress/eval/gen_paper_tables.py -- do not hand-edit.

\begin{table}[t]
\centering
\small
\setlength{\tabcolsep}{5pt}
\caption{Equal-memory serving at $\budget{=}1024$ (Qwen3-4B, $32$k generations, one $143$\,GB H200): the tighter budget fits a batch of $584$ against $28$ for full attention, and \method{} reaches $28.8\times$ full-attention throughput. \rkv{} and \triattn{} were not measured at this budget.}
\label{tab:isomem_k1024}
\begin{tabular}{lccc}
\toprule
 & \multicolumn{3}{c}{Qwen3-4B} \\
\cmidrule(lr){2-4}
Method & batch & tok/s & $\times$\,full \\
\midrule
\dense{} (no eviction) & 28 & 178 & 1.00 \\
\snapkv{} & 544 & 4400 & 24.79 \\
\vase{} & 552 & 4273 & 24.07 \\
\rowcolor{oursbg} \method{} (ours) & 584 & \best{5110} & \best{28.79} \\
\bottomrule
\end{tabular}
\end{table}

\paragraph{Equal memory at a tighter budget.} The multiple over full
attention is set by the batch the cache admits, so it grows as the
budget shrinks. Table~\ref{tab:isomem_k1024} repeats the protocol on
Qwen3-4B at $\budget{=}1024$, the \mathfive{} budget of
Table~\ref{tab:main}: the compressed caches now fit $544$--$584$
sequences against $28$, and \method{} serves $28.8\times$ the
full-attention throughput, $16\%$ more than \snapkv{} and $20\%$ more
than \vase{} at their own largest batches. 

\paragraph{Quoting \triattn{}'s \rkv{} comparison.} We cite their
matched-budget row rather than their headline. Their reported throughput
varies almost entirely with the budget and hardly at all with the selector
($1405$, $760$, $564$ and $414$ tok/s at $\budget{=}1024$, $2048$, $3072$
and $4096$), and the headline $+85\%$ over \rkv{} places the two methods
at different budgets ($1024$ for \triattn{}, $2048$ for \rkv{}), so it
measures the budget, not the selection signal. At equal budget their
table reports $1405.2$ against $1345.5$, which is the $+4.4\%$ we quote and
the conservative figure for our purposes. That row was measured on an A100
at $16$k decode and maximum batch, neither our hardware nor our operating
point, so it combines with our own measurement only as an indication, not as
a controlled comparison.

\paragraph{What porting an evictor to a paged runtime involves.} Two
settings are necessary because no released runtime hosts every method.
\triattn{} ships a vLLM v0.19.0 plugin, which is what Table~\ref{tab:vllm}
uses. \rkv{} ships ports as well, but against a different pinned vLLM
release, so the two cannot serve in one comparison without re-porting one of
them; \snapkv{} and \vase{} are released as HuggingFace-side implementations.
The wiring is the larger part of the work and is independent of the score:
physical eviction under paging has to rewrite each request's surviving KV
into its blocks, free the tail, keep rotary positions logical while the
physical slots shrink, and stay correct across preemption; \rkv{}'s
released port spans ${\sim}849$ lines of wiring across $13$ upstream files
and requires vLLM's V1 model runner, and the integration we serve on mishandles
preemption (above). The score then decides what else is needed. A score read
from the cache alone (\vase{}'s value range, \triattn{}'s calibrated key
statistics) costs gathers across the block table. A score read from
attention weights (\snapkv{}, \rkv{}'s importance term, \vase{}'s
attention-proportional fill) cannot be read at all from a fused paged
kernel, which never materialises the weights, so it must either be
recomputed as an explicit window-query product against every candidate key
(chunked, since the transient is $\text{heads} \times \text{window} \times
\text{cache}$ per request) or be extracted by modifying the attention kernel
itself. \method{} needs neither: its keep-set is a random permutation of slot
indices, so it reads nothing and the runtime's existing compaction path is
the whole integration, which is why adding it to \triattn{}'s plugin
took a single function.

\paragraph{Why the two settings report such different multiples.} On
Qwen3-14B at $32$k generations, \method{} runs at $1436$ tok/s here and
$1819$ in vLLM, while full attention runs at $164$ here and $925$ there: the
compressed method gains ${\sim}1.3\times$ from the better engine and the full-attention
baseline gains ${\sim}5.6\times$. Paging is the reason for the asymmetry:
reserving a full cache per sequence is exactly the constraint compression
relieves, and a paged allocator relieves it too, admitting requests as
memory frees rather than capping the batch at $20$. The compressed methods have
little left to gain, since their caches are small under either allocator.
The two settings also differ in budget ($3072$ here, $2048$ there) and in
request count, so this attribution is indicative rather than a controlled
decomposition; what it explains is why an unpaged capacity multiple of
$8.8\times$ and a paged one of $1.97\times$ are consistent measurements of
the same effect.

\section{Discussion and limitations}
\label{sec:discussion}

\paragraph{What the evidence covers.} Our experiments study decode-phase eviction for long chain-of-thought reasoning: prompts of a few hundred tokens, traces of several thousand to $32$k tokens, budgets from $2\times$ to $16\times$ compression, four reasoning models, six tasks in math, science and code, and training-free evictors as baselines. The claim that the selection signal contributes almost nothing is a claim about this regime, and ``strongest baseline'' means the strongest of the training-free methods we evaluate. It is also a claim about the aggregate rather than every cell: four of the $80$ baseline comparisons in Tables~\ref{tab:main} and~\ref{tab:appendix_grid} favour a baseline significantly (\triattn{} on \lcb{} at 14B and 32B and on \mathfive{} at 14B, and \vase{} on \aime{} at 14B). Long-input workloads, where the prompt itself fills the cache, fall outside this regime and have their own evidence \citep{protectiondominates}.

\paragraph{What the claim rests on.} The $31$ significant wins in the main grid are against the baselines as released, and \S\ref{sec:mech_protect} shows that most of them come from prompts those baselines lose; they establish that \method{} is competitive, not that scores carry no information. The evidence about the score itself comes from the comparisons in which protection is matched. Against \triattn{}, which keeps the prompt as \method{} does, the $20$ cell differences in Tables~\ref{tab:main} and~\ref{tab:appendix_grid} range from $-2.8$ to $+2.9$ points and average $0.0$. With the prompt given to every other baseline, none of the $27$ protected baseline cells of Tables~\ref{tab:promptprotect} and~\ref{tab:promptprotect_appx} scores above \method{}. We state this parity descriptively rather than through a formal equivalence test, and a non-significant cell is not by itself evidence of equality; the claim rests on the pattern across these matched cells, not on any one of them.

\paragraph{Signal-free, not structure-free.} \method{} computes no score, but it is not free of structure. It pins the prompt, and because every surviving position faces a fresh draw at each eviction, the survival of a trace position decays geometrically with its age, about $0.94^n$ after $n$ evictions at $\budget{=}1024$ (Appendix~\ref{sec:keeplog}); the policy behaves as a soft recency window. This age profile is not a design choice: any memoryless draw produces it, and at each eviction every trace position is treated alike. By ``selection signal'' we mean what the baselines add on top of such structure, a score computed from the cache (attention weights, values or keys). Our claim is that this score adds little once the prompt is kept, not that structure is unimportant; protection is the largest effect we measure.

\paragraph{Protected recency.} The nearest signal-free alternative is therefore a hard recency window with the prompt kept. In the four matched settings of Table~\ref{tab:promptprotect} it is never more than two points below the best protected baseline, and \method{} is above it in all four, by $1.1$--$3.1$ points. The two spend the same budget differently: the window keeps every recent position and nothing older than the budget, while \method{} keeps a thinning sample of both and still holds $19\%$ of the $1$--$2$k band in each head and $78\%$ across heads at $\budget{=}1024$ (Appendix~\ref{sec:keeplog}). The tail can help only when the model needs something older than the budget that it has not restated within it, and it cannot help once too few copies survive, as in the passcode test of \S\ref{sec:mech_boundary}. We ran the protected window only in these diagnostic settings, not across the full grid or the compression sweep; the two policies cost the same, and where we ran both, \method{} scored higher. We also centre the paper on \method{} because a null for a selection signal should not assume which trace positions matter, and a window assumes that only recent ones do. Were a protected window to match \method{} everywhere, our main conclusion would stand, since neither computes a score.

\paragraph{Relation to other protection results.} This conclusion agrees with \citet{protectiondominates}, who finds that protection dominates scoring for long-context question answering under a global cache cap, and with the concurrent Prefix Sliding \citep{muennighoff2026prefix}, which uses protected recency during reasoning; what we add is matched evidence in decode-phase reasoning against four scored evictors, an account of why the trace tolerates signal-free eviction, and the serving cost of scoring.

\paragraph{Matched protection across the grid.} The main grid runs every baseline as released, because that is what a practitioner would deploy, so each cell mixes the effect of the score with whatever protection the method applies by default. We separate the two where the grid showed large, method-specific gaps: the four settings of Table~\ref{tab:promptprotect} and the code, competition-math and 32B \gpqa{} settings of Appendix~\ref{sec:ppappendix}; every \triattn{}-versus-\method{} cell is already matched, since both keep the prompt. We have not run the protected grid on Qwen3-14B, but the missing cells would not explain away the three baseline wins there: two are \triattn{}'s, already matched, and \vase{}'s \aime{} win is on competition math, where the rule did not help \vase{} on Qwen3-4B ($-2.1$ and $-1.0$ points on \aime{} and \hmmt{}, Table~\ref{tab:promptprotect_appx}). What the missing cells would change is the size of \method{}'s margins over \snapkv{}, \rkv{} and \vase{}, which the rule narrows where a baseline had been losing the prompt.

\paragraph{Which redundancy does the work.} The two levels of redundancy have different evidential status. That copies of a fact pool across heads is shown directly, but only in the planted-fact probe, where the text is not redundant by construction. On real \mathfive{} traces a shared draw, which keeps the same positions in every head, scores within $0.3$ points of \method{} at $4\times$ and $8\times$ compression (Appendix~\ref{sec:keeplog}). On the traces we tested, then, the text carries the load, and cross-head redundancy is a second line of defence for what the text does not restate; we present the probe as evidence of that mechanism, not as the cause of the benchmark results. Text-level redundancy is itself inferred, not measured: it is documented for reasoning traces by \citet{cai2025rkv}, and it is what reconciles our two observations that random eviction preserves benchmark accuracy while a once-stated passcode is lost. Independent per-head draws cost nothing over a shared draw, so we keep them, but we have run the shared draw only on \mathfive{} and the probe only on Qwen3-4B.

\paragraph{Where a selection signal still pays.} The clearest case is a fact stated once, never restated, and needed much later: \method{} never recovers a passcode announced $57$ compression rounds before the question, while \rkv{} recovers it $84\%$ of the time (Table~\ref{tab:synthboundary}). Workloads built around such facts, for example state that an agent reads once and consults much later, are where a content-dependent signal earns its cost, and \method{} is the wrong default there. We have not measured how often real reasoning traces contain such facts, that is, how often a value the model uses was last stated more than a budget earlier; the aggregate results suggest they are rare on our benchmarks. A natural hybrid keeps the random draw for most of the budget and reserves a few slots per head for a cheap content signal, much as \vase{} reserves slots for large values; we have not evaluated one.

\paragraph{The code gap.} Code is the only task on which a content signal wins at more than one scale: \triattn{} leads on \lcb{} at 14B and 32B, by $2.3$ and $2.8$ points. Both methods pin the prompt, so the prompt does not produce this gap by itself; what it does is shrink the budget left for the trace, to as little as half of $\budget{}$ on the longest code prompts. In the compression sweep on Qwen3-4B and Phi-4-reasoning, far smaller trace budgets on the math and science tasks leave \method{} within a few points of \triattn{} or ahead of it (Figure~\ref{fig:regime}), so a small trace budget alone does not open a gap there. Code traces that define names and interfaces once and use them much later would sit closer to the once-stated regime above, which would explain the gap; we have not run the prompt-length-stratified analysis, or the sweep at the larger models, that would test this.

\paragraph{Budgets and long prompts.} \method{} spends part of its budget on the prompt before any selection happens, and this is its main cost. On \lcb{}, the longest prompts take up to half of the $\budget{=}3072$ budget, and at $16\times$ compression on \mathfive{} ($\budget{=}256$) the protected prompt no longer fits for both prompt-keeping methods on Phi-4-reasoning and for \triattn{} on Qwen3-4B (the $\times$ markers in Figure~\ref{fig:regime}); \vase{}, which scores the prompt, degrades more gracefully there. % CHECK: adjust if the TriAttention x on Qwen3-4B has another cause
The method is therefore defined for budgets that hold the prompt. Long-context and multi-turn settings, where the prompt or conversation history can exceed any practical budget, need a rule for which parts to protect, and compacting the scaffolding of long code prompts is a smaller version of the same question; we leave both to future work.

\paragraph{Architectures.} All four models use grouped-query attention with eight or ten key-value heads per layer, and three of them are Qwen3 sizes; Phi-4-reasoning differs in family, tokenizer and training data. We have not tested architectures where per-head independence is unavailable or different in kind: multi-head latent attention, whose heads share one latent cache per layer, multi-query attention, or hybrids of full and sliding-window layers. The shared-draw result suggests that text-level redundancy alone carries real traces at our budgets, which is the case that matters when per-head draws are unavailable, but this is an inference, not a measurement.

\paragraph{Efficiency.} The paged-serving comparison includes only \triattn{}, the one baseline whose released integration runs on the vLLM version we use (Appendix~\ref{sec:effdetail}), and its $32$--$43\%$ margin is a property of that integration as much as of the algorithm. The score's own cost is small: about two percent of single-stream decoding time (Table~\ref{tab:evictcost}), and about one percent end to end at a single request. The integration multiplies it by compressing at a synchronisation point between batched steps, so the margin grows with load, from $30$--$35\%$ at $64$ concurrent requests to $41$--$42\%$ at capacity; a fused or asynchronous scorer could remove much of this amplification, though not the score's own cost, which \method{} does not pay. Compression itself pays only when the cache limits concurrency: at $8$k-token generations every compressed method serves less than full attention (\method{} $0.52$--$0.96\times$, \triattn{} $0.38$--$0.70\times$), while \method{}'s lead over \triattn{} holds. The throughput runs use fixed output lengths, whereas under natural stopping the methods generate different amounts: averaged over the five task columns, \method{} generates fewer tokens than \triattn{} on three of the four models and $5\%$ more on Qwen3-32B (Appendix~\ref{sec:lensvar}), well inside its throughput margin. The exception is \lcb{} on Qwen3-32B, where \method{}'s traces are $44\%$ longer and its end-to-end advantage would not hold.

\paragraph{Statistics and randomness.} Every margin we call significant passes both a paired, problem-clustered bootstrap and an exact sign test at the $5\%$ level, each comparison on its own, without a correction for multiple comparisons. Over $80$ comparisons a few cells in either direction may be significant by chance, including the weakest of the baseline wins ($p{=}.02$, Appendix~\ref{sec:generality}), which is why no conclusion rests on a single cell. Eviction randomness adds to the randomness of sampled decoding; every accuracy averages $2$ to $16$ independently sampled runs, whose spread Appendix~\ref{sec:lensvar} reports, and a per-request seed makes \method{} reproducible when needed. % CHECK: App A says the Random Attention AIME cell pools 32; if so, write "2 to 32"

\paragraph{Learned evictors.} Learned evictors such as TrimKV \citep{bui2026cache}, DBTrimKV \citep{bui2026make} and ForesightKV \citep{dong2026foresightkv} require training a scorer for each base model on domain-specific data, so we compare only against training-free evictors, which run on any model as released. They relate to our findings in two ways: TrimKV and DBTrimKV learn how fast each token's score decays with age, a learned, per-token version of the uniform rate at which a position's survival decays under \method{} (Appendix~\ref{sec:keeplog}), while ForesightKV learns a content score from a future-attention oracle, the kind of content-dependent signal that recovered the once-stated passcode in \S\ref{sec:mech_boundary}. 

\end{document}